\documentclass[runningheads]{llncs}

\usepackage{graphicx}
\usepackage{amsmath,amssymb} % define this before the line numbering.
\usepackage{color}
\usepackage{subcaption}
\usepackage{times}
\usepackage{epsfig}
\usepackage{bm}
\usepackage[export]{adjustbox}
\usepackage{float}
\usepackage{lineno}

\newcommand{\ignore}[1]{}

\newcommand{\warp}{\omega}

\def\setof#1{{\left\{#1\right\}}}

\def\calC{\mathcal{C}}

\def\calL{\mathcal{L}}
\def\calM{\mathcal{M}}
\def\calN{\mathcal{N}}

\newcommand{\maybesubfigure}[3]{
	\IfFileExists{#1}{
		\begin{subfigure}[b]{0.31\textwidth}
			\setlength{\abovecaptionskip}{1pt}
			\setlength{\belowcaptionskip}{8pt}
			\includegraphics[width=\textwidth]{#1}
			\caption{\label{fig:#3}#2}
		\end{subfigure}
	}{}}

\newcommand{\maybeimage}[2]{
	\IfFileExists{#1}{\includegraphics[width=#2\textwidth]{#1}}{(No Result)}
}

\usepackage[pagebackref=true,breaklinks=true,colorlinks,bookmarks=false]{hyperref}

\usepackage{cleveref}

\begin{document}
	
	%%%%%%%%% TITLE
	\pagestyle{headings}
	\mainmatter
	\def\ECCV18SubNumber{1519}  % Insert your submission number here
	
	\title{
		Object-centered image stitching
	}
	
	\titlerunning{Object-centered image stitching}
	
	\authorrunning{C. Herrmann, C. Wang, R.S. Bowen, E. Keyder and R. Zabih}
	
	\author{Charles Herrmann\inst{1} \and Chen Wang\inst{1,2} \and Richard Strong Bowen\inst{1} \and\\ Emil Keyder\inst{2} \and Ramin Zabih\inst{1,2}}
	\institute{
		Cornell Tech, New York, NY 10044, USA \and
		Google Research, New York, NY 10011, USA \\
		\email{\{cih,chenwang,rsb,rdz\}@cs.cornell.edu, \{wangch,emilkeyder,raminz\}@google.com}
	}
	% To Springer editor: in "Richard Strong Bowen", Richard is the first name, Strong is the middle name and Bowen is the last name.

	\maketitle

	\begin{abstract}
		Image stitching is typically decomposed into three phases:
		registration, which aligns the source images with a common target image;
		seam finding, which determines for each target pixel the source image it
		should come from; and blending, which smooths transitions over the seams.  As described in \cite{Szeliski:tutorial:2006}, the seam
		finding phase attempts to place seams between pixels where the transition
		between source images is not noticeable. Here, we observe that the most
		problematic failures of this approach occur when objects are cropped,
		omitted, or duplicated. We therefore take an object-centered approach to the
		problem, leveraging recent advances in object detection
		\cite{Goodfellow16,Liu:SSD16,Ren:RCNN:PAMI17}. We penalize candidate solutions with this class of error by modifying the energy function used in the seam finding
		stage. This produces substantially more realistic stitching results on
		challenging imagery. In addition, these methods can be used to
		determine when there is non-recoverable occlusion in the input data, 
		and also suggest a simple evaluation metric that can be used to 
		evaluate the output of stitching algorithms.
	\end{abstract}
	
	\section{Image stitching and object detection}
	\label{sec:introduction}
	
\begin{figure}[t]
		\centering
		\begin{subfigure}[c]{0.32\textwidth}
			\centering
			\includegraphics[width=\textwidth]{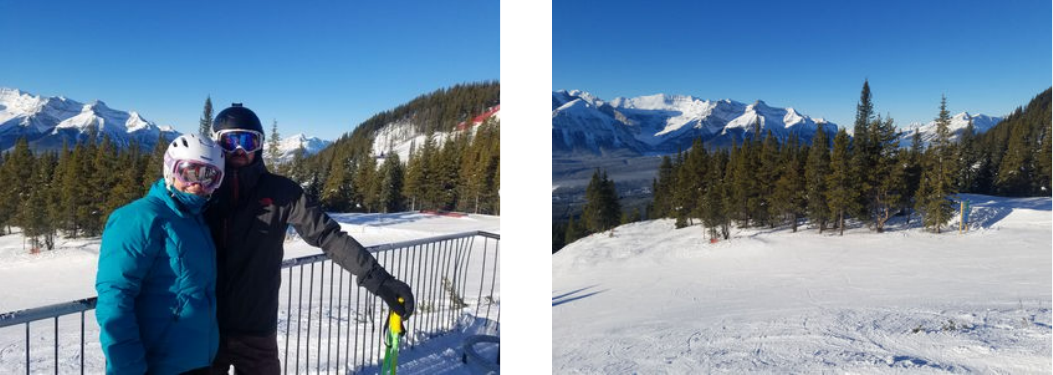}
			\caption{Input images}
		\end{subfigure}
		\begin{subfigure}[c]{0.32\textwidth}
			\centering
			\includegraphics[width=\textwidth]{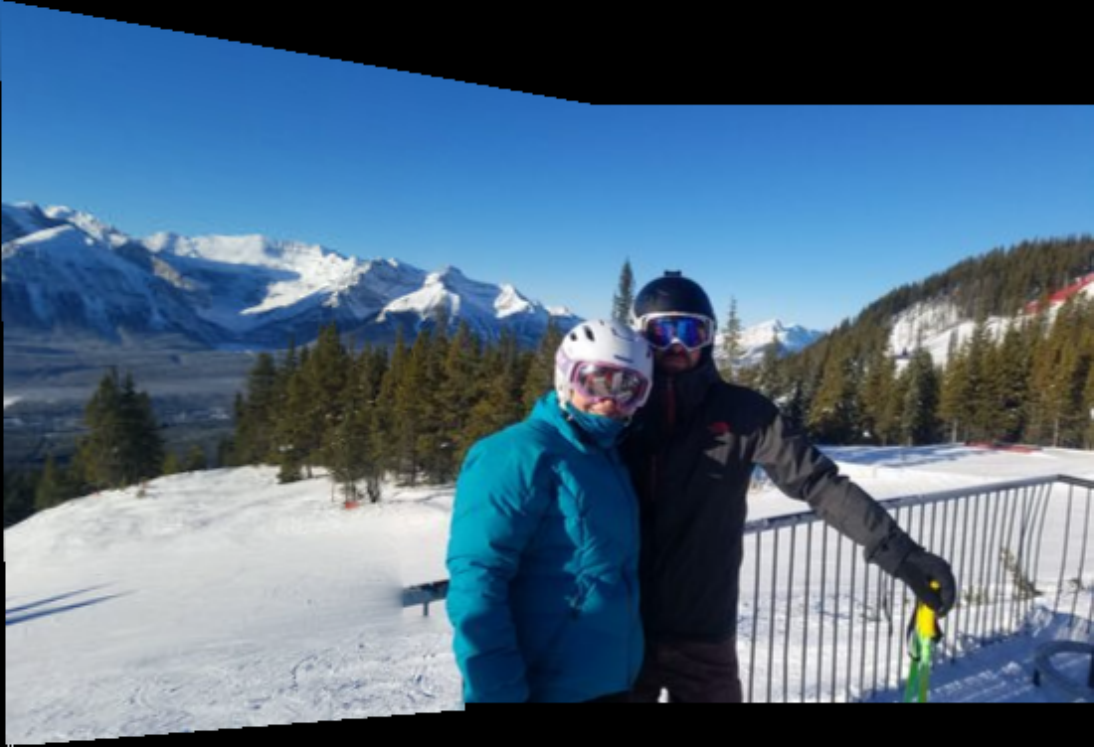}
			\caption{Our results}
		\end{subfigure}
		\begin{subfigure}[c]{0.32\textwidth}
			\centering
			\includegraphics[width=\textwidth, height=75px]{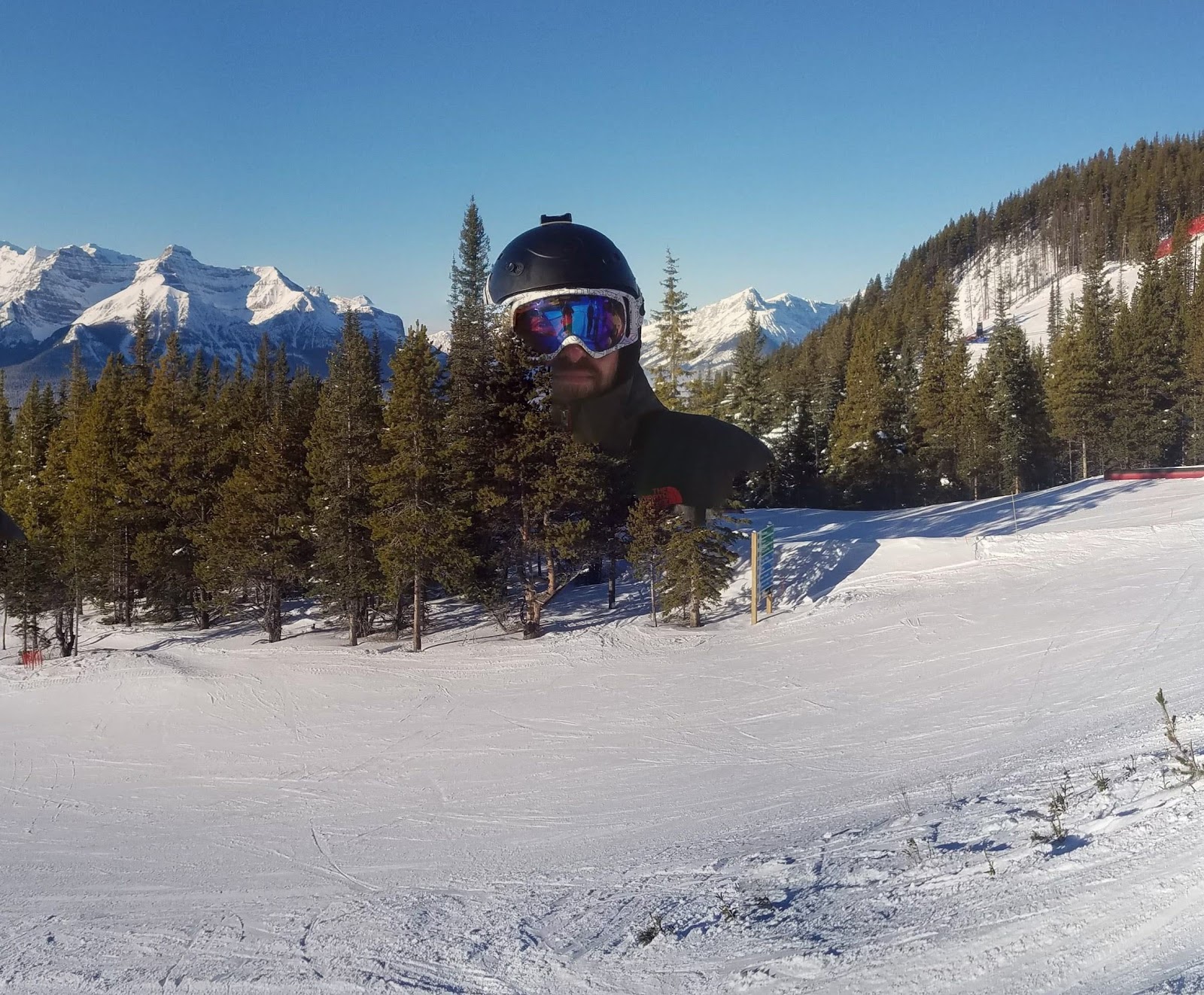}
			\caption{Google Photos results}
		\end{subfigure}
		\caption{Example showing object cropping. Google Photos (shown) crops the man's body and blends him into the mountains. APAP \cite{zaragoza2013projective} and 
			Adobe Photoshop both only include the man's arm, while NIS \cite{Chen2016} produces severe ghosting.
			\label{fig:ski-motivation}
		}
	\end{figure}

\begin{figure}[t]
		\centering
		\begin{subfigure}[c]{0.32\textwidth}
			\centering
			\includegraphics[width=\textwidth]{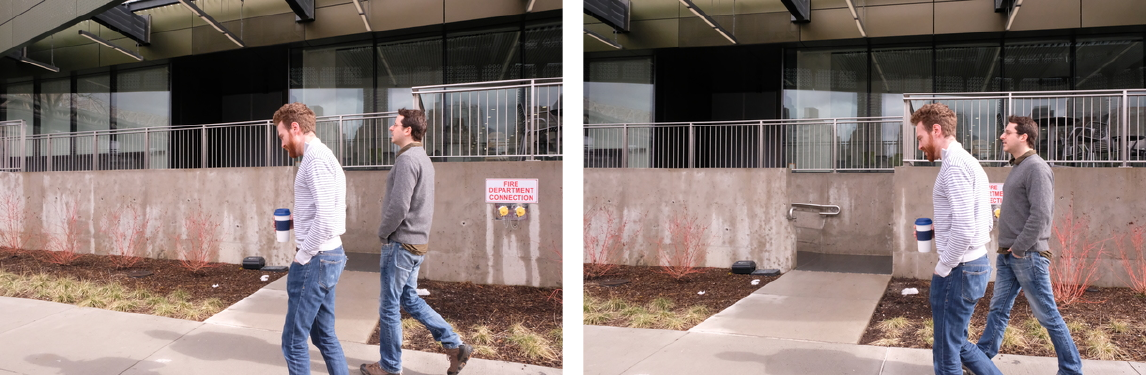}
			\caption{Input images}
		\end{subfigure}
		\begin{subfigure}[c]{0.32\textwidth}
			\centering
			\includegraphics[width=\textwidth]{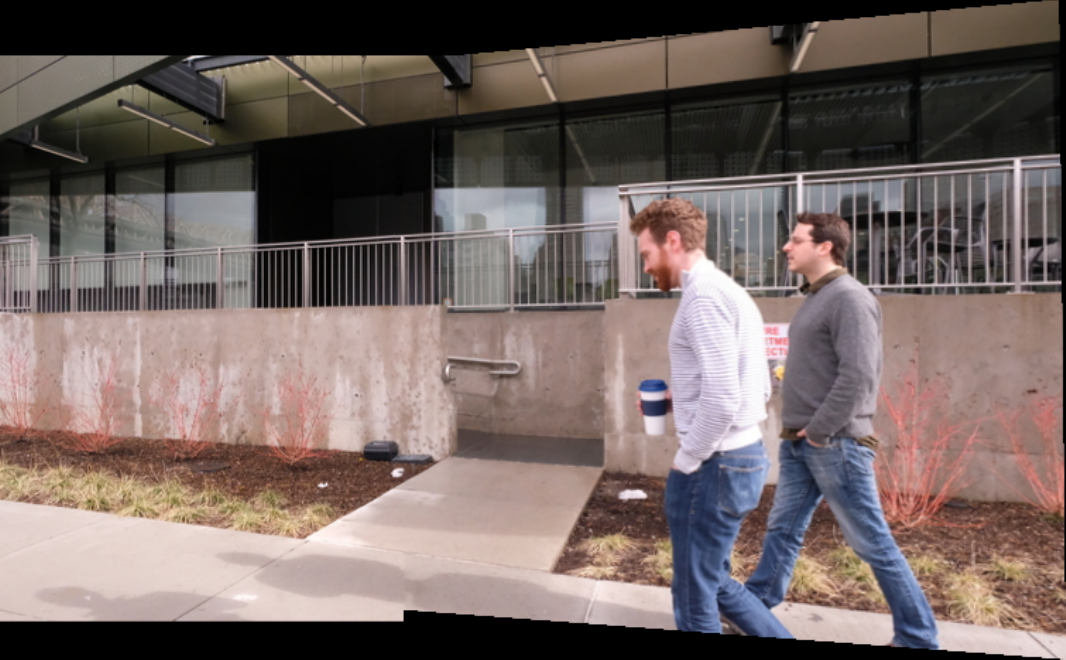}
			\caption{Our results}
		\end{subfigure}
		\begin{subfigure}[c]{0.32\textwidth}
			\centering
			\includegraphics[width=\textwidth]{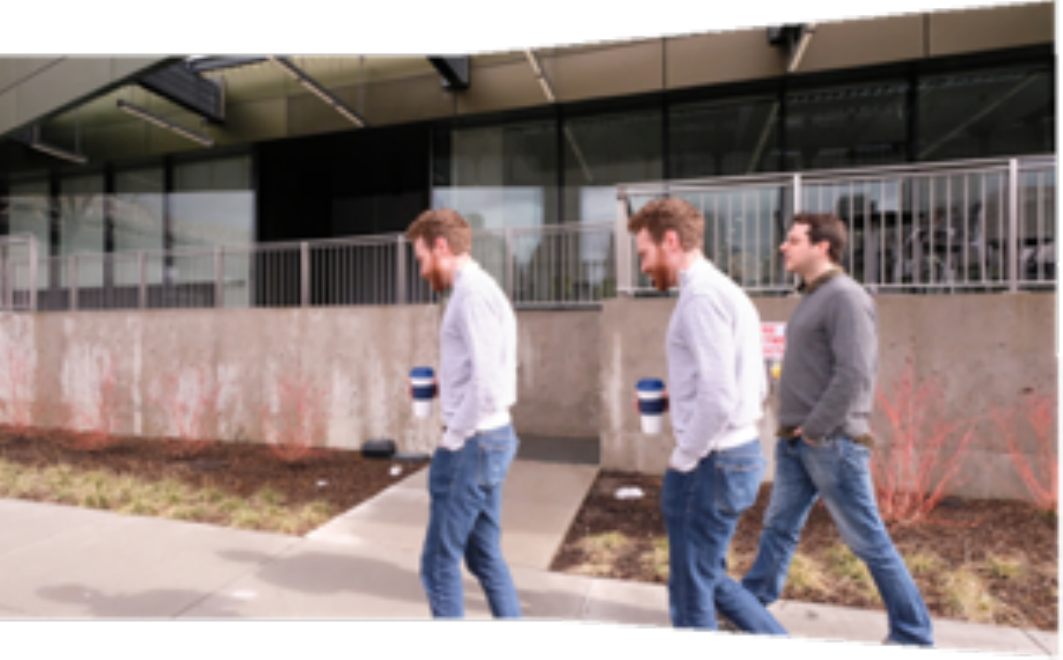}
			\caption{Photoshop results}
		\end{subfigure}
		\caption{Example showing object duplication. Photoshop (shown) and APAP
			give visually similar output, while NIS produces severe ghosting. While we duplicate a shadow on the sidewalk, our object-centered approach preserves
			the most important elements of the scene.
			\label{fig:richards-motivation}
		}
	\end{figure}

	Image stitching is the creation of a single composite image from a set
	of images of the same scene. It is a well-studied problem
	\cite{Szeliski10} with many uses in both industry and consumer applications,
	including Google StreetView, satellite mapping, and the panorama creation software 
	found in modern cameras and smartphones. Despite its ubiquitous  applications, 
	image stitching cannot be considered solved. Algorithms frequently produce images that appear 
	obviously unrealistic in the presence of parallax (Figure~\ref{fig:ski-motivation}(c)) or object motion (Figure~\ref{fig:richards-motivation}(c)), or alternatively indicate that images are too disparate to be 
	stitched when this is not the case. One of the most visually jarring failure modes is the tearing, cropping, 
	deletion, or duplication of recognizable objects. Indeed, it has become a popular internet
	pastime to post and critique stitching failures of this sort that occur in Google StreetView  or on users' own cameras (most famously, the Google Photos failure shown in 
	Figure~\ref{fig:ski-motivation}). 
	In this paper, we exploit advances in object 
	detection \cite{Goodfellow16,Liu:SSD16,Ren:RCNN:PAMI17} to improve image stitching algorithms 
	and avoid producing these artifacts.
	
	The image stitching pipeline typically consists of three phases: {registration}, 
	in which the images to be stitched are
	aligned to one another; {seam finding}, in which a source image is
	selected for each pixel in the final image; and {blending}, in which smooths over the transitions between images \cite{Szeliski10}. In order to avoid introducing
	object-related errors, we propose modifications to the seam finding step,
	which typically relies on Markov Random Field (MRF) inference
	\cite{Szeliski10,Kwatra:2003}.  We demonstrate that MRF inference can be
	naturally extended to prevent the duplication and maintain the integrity of detected
	objects.  In order to evaluate the efficacy of this approach, we experiment
	with several object detectors on various sets of images, and
	show that it can substantially improve the perceived quality of the
	stitching output when objects are found in the inputs.\footnote{In the atypical case of no detected objects, our technique reverts to standard stitching. As object detectors continue to improve their accuracy and coverage, this situation will likely become exceptionally rare.} We also show that object 
	detection algorithms can be used to formalize the evaluation of stitching results, 
	improving on previous evaluation techniques \cite{Li:SIVP2018} that require knowledge of seam
	locations.
	
	In the remainder of this section, we give a formal description of the stitching 
	problem, and summarize how our approach fits into this framework.
	Section~\ref{sec:related} gives a short review of related work. In
	Section~\ref{sec:object-approach}, we present our object-centered approach for
	improving seam finding. We propose an object-centered evaluation metric for
	image stitching algorithms in Section~\ref{sec:eval-stitch}. Section~\ref{sec:data}
	gives an experimental evaluation of our techniques, and Section~\ref{sec:limits} 
	discusses their limitations and possible extensions.

	%% In \cite{Paper1}a parallel submission, we take a very different approach to
	%% improving stitching algorithms, relying on multiple registrations of a
	%% pair of images instead of object detection. Following the ECCV18
	%% submission instructions, we have included that submission in the
	%% supplemental material.  In this submission, we use a
	%% technique for computing a single non-rigid registration that is
	%% adapted from \cite{Paper1}.  An explicit comparison against the
	%% experimental results obtained there is included below in
	%% Section~\ref{sec:limits}.
	
	%\subsection{Motivating examples}
	
	% \begin{figure}[t]
	% \hspace*{-.1in}
	% \includegraphics[height=2.75in]{P2heros_ski.pdf}
	% \vspace*{-1.5in}
	% \caption{Example showing object cropping. Google Photos (shown). APAP \cite{zaragoza2013projective} and 
	% Adobe Photoshop both only include the man's arm, while NIS \cite{Chen:NIS:ECCV16} produces severe ghosting.
	%   \label{fig:ski-motivation}
	% }
	% \end{figure}
	% \begin{figure}
	% \hspace*{-.1in}
	% \includegraphics[height=2.75in]{P2heros_richards.pdf}
	% \vspace*{-1.5in}
	% \caption{Example showing object duplication. Photomerge (shown) and APAP
	%  give visually similar output, while NIS produces severe ghosting. Note that
	%  we duplicate a shadow on the sidewalk but due to object detection preserve
	%  the most important elements of the scene.
	% \label{fig:richards-motivation}
	% }
	% \end{figure}
	
	\subsection{Formulating the stitching problem}
	
	We use the notation from \cite{Paper1} and
	formalize the perspective stitching problem as follows: given two images\footnote{we address the generalization to
		additional overlapping images shortly}
	$I_1,I_2$ with an overlap, compute a
	registration $\warp(I_2)$ of $I_2$ with respect to $I_1$, and a \emph{label} $x_p$ 
	for each pixel $p$ that determines whether it gets its value from $I_1$ or from $\warp(I_2)$.
	
	Following \cite{Kwatra:2003}, the label selection problem is typically solved with an MRF that uses an energy function that prefers short seams 
	with inconspicuous transitions between $I_1$ and $\warp(I_2)$. The energy to be 
	minimized is
	
	\begin{equation*}
	E(x) = \underset{x \in \calL}{\arg\min} \sum_p E_d(x_p) \lambda_d [M_i(p) = 0] + \sum_{p,q\in \calN} V_{p,q}\cdot [x_p \neq x_q].    
	\end{equation*}
	
	The underlying data term $E_d$ is combined with a factor $\lambda_d [M_i(p) = 0]$, where $[]$ are Iverson
	brackets and $M_i$ is a \emph{mask} for each input $i$ that has value 1 if image $I_i$ has a value at that
	location and 0 otherwise. This guarantees that pixels in the output are preferentially drawn from valid regions
	of the input images.
	
	For a pair of adjacent pixels $p,q \in \calN$, the prior term $V_{p,q}$ imposes a penalty
	for assigning them different labels when the two images have different
	intensities. A typical choice is
	\begin{math}
	V_{p,q} = |I_1(p) - \warp(I_2)(p)| + |I_1(q) - \warp(I_2)(q)|.  
	\end{math}
	
	%% The standard MRF seam-finding technique \cite{Kwatra:2003}
	%% minimizes the energy defined by the two images $I_1,\warp(I_2)$:
	
	%% We then create the composite image by choosing for each pixel $p$ the
	%% intensity of one of $I_1,\warp(I_2)$, depending on the value of the label
	%% $x_p$.
	
	The generalization to multiple overlapping images is straightforward: 
	with a reference image $I_1$ and $k-1$ warped images
	$\setof{\warp_2(I_2),\warp_3(I_3),\ldots,\warp_k(I_k)}$, the size of
	the label set is $k$ instead of $2$
	%, one for $I_1$ and one for each
	% warped image.
	and the worst-case computational complexity goes from polynomial to NP-hard \cite{BVZ:PAMI01}. Despite this theoretical complexity, modern MRF inference methods such as graph cuts are very effective at solving these problems \cite{SZSVKATR:PAMI08}.

	\textbf{Our approach.} We focus primarily on 
	modifications to the seam finding stage. We introduce three new terms to the 
	traditional energy function that address the cropping, duplication, and occlusion of objects.
	We also demonstrate that object detection can be used to detect cropping and 
	duplication on the outputs of arbitrary stitching algorithms.
	
	\section{Related work}
	\label{sec:related}
	
	A long-standing problem in image stitching is the presence of visible seams
	due to effects such as parallax or movement. Traditionally there have been two ways of
	mitigating these artifacts: to improve registration by increasing
	the available degrees of freedom \cite{Chen2016,Lin_2015_CVPR,Lin2016}, or to hide misalignments by selecting better seams. We note that artifacts caused by movement \emph{within} the scene cannot be concealed by better registration, and that improved seams are the only remedy in these cases.
	
	%% PUT SOMETHING HERE ABOUT HOW ALIGNMENT CANNOT SOLVE FOR MOVEMENT?? Done - Emil
	
	%% In this
	%% paper, we focus on the seam finding portion of the problem. Alignment techniques often struggle %% with scenes with large
	%% amounts of irregular movement and do not 
	
	%% \subsection{Registration}
	%% \label{subsec:related-registration}
	
	%\input{related_work_flow}
	
	%% \subsection{Seam Finding}
	%% \label{subsec:related-seam-finding}
	
	Our work can be seen as continuing the second line of research. Initial approaches here based the pairwise energy term purely on differences in intensity between
	the reference image and the warped candidate image
	\cite{Kwatra:2003}. This was later improved upon by considering
	global structure such as color gradients and the presence of
	edges\cite{Agarwala:2004:IDP:1015706.1015718}.
	
	A number of papers make use of semantic information in
	order to penalize seams that cut through entities that human observers are especially likely to notice, such as faces \cite{Ozawa:2012:HPI:2160125.2160145}. One more general approach modifies the energy function based on a \emph{saliency} measure defined in terms of the location in the output image and human
	perceptual properties of colors \cite{Li:SIVP2018}. Our methods differ from these in
	that we propose general modifications to the energy function that
	also cleanly handle occlusion and duplication. \cite{Serge_2010} uses graphcuts to remove
	pedestrians from Google StreetView images; their technique bears a strong similarity to our duplication term but addresses
	a different task.
	
	Evaluation of image stitching methods is very difficult, and has been a major
	roadblock in the past. Most MRF-based stitching methods report
	the final energy as a measure of quality
	\cite{Kwatra:2003,SZSVKATR:PAMI08}, and therefore cannot be used to compare
	approaches with different energy functions, or non-MRF based methods. \cite{Li:SIVP2018} proposes
	an alternate way to evaluate stitching techniques based on seam quality; their work is
	perceptually based but similar to MRF energy-based approaches. Our approach, in contrast, takes
	advantage of more global information provided by object detection.
	
	%% \subsection{Comparisons with our technique}
	%% \label{subsec:related-comparisons}
	
	% \input{model}
	
	%\input{math}
	
	\section{Object-centered seam finding}
	\label{sec:object-approach}
	
	We use a classic three-stage image stitching pipeline, composed of
	registration, seam finding, and blending phases. We modify the techniques
	introduced in \cite{Paper1} to find a single best registration of the
	candidate image. We then solve a MRF whose energy function incorporates our novel
	tearing, duplication, and occlusion terms to find seams between the images. 
	Finally, we apply Poisson blending \cite{Perez:Poisson2003} to smooth transitions 
	over stitching boundaries to obtain the final result.
	
	\subsection{Registration}
	
	Our registration approach largely follows \cite{Paper1}, except that we only
	use a single registration in the seam finding stage. To generate a
	registration, we first identify a homography that matches a large portion of
	the image and then run a content-preserving warp (CPW) in order to fix small
	misalignments~\cite{Liu:SIGGRAPH09}. The following provides a high level
	overview of the registration process.
	
	To create candidate homographies, we run RANSAC on the sparse
	correspondence between the two input images. In order to limit the set
	of candidates, homographies that are too different from a similarity 
	transform or too similar to a previously considered one are filtered out
	at each iteration. The resulting homographies are then refined via 
	CPWs by solving a quadratic program (QP) for each, 
	in which the local terms are populated from the results of an optical 
	flow algorithm run on the reference image and initial candidate registration. 
	This step makes minor non-linear adjustments to the transformation and yields registrations 
	that more closely match portions of the image that would otherwise be 
	slightly misaligned.
	
	%% Given the homography $H$ that is
	%% output by RANSAC, we solve a quadratic program (QP) that modifies $H$
	%% to minimize the sum of smoothed correspondence distance; this makes
	%% minor adjustments to the homography and enables it to capture a larger
	%% number of points more evenly. 
	
	%% After the homography phase, we run an optical flow algorithm on the
	%% reference image and candidate registration. The resulting optical flow
	%% is used to populate the local term in a CPW, where the local term
	%% expresses the desired motion for each pixel. This applies a non-rigid
	%% transformation to the homography and allows us to obtain registrations
	%% that more closely match portions of the image that would otherwise be
	%% slightly misaligned.
	
	We also explored producing multiple registrations and running seam
	finding on each pair of reference image and candidate registration, and 
	selecting the result by considering both the lowest final energy
	obtained and the best evaluation score (defined below). This approach
	is similar to the process used in \cite{Lin2016,Zhang_2014_CVPR} where
	homographies are selected based on the final energy from the seam
	finding phase. We found that this method gave only marginally better 
	results than selecting a single registration.
	
	\subsection{Seam finding}
	
	The output of the registration stage is a single proposed warp
	$\warp(I_2)$.  For simplicity, let $I^S_1 = I_1$, $I^S_2 = \warp(I_2)$
	be the input images for the seam finding phase. We denote the set of pixels in the output mosaic by $P$. In contrast to the traditional seam finding setup, here we assume an
	additional input consisting of the results of an object
	detector run on the input images. We write the set of recognized
	objects in $I^S_\ell$ as $O_\ell$  \ignore{Each object $o\in
		O_\ell$ is represented by the set of pixels in that object.}  and
	denote by $\calM(O_1, O_2) \subseteq O_1 \times O_2$ the set of
	corresponding objects between $O_1$ and $O_2$.  The computation of  $\calM(O_1, O_2)$ is discussed in Section~\ref{subsec:establishing-correspondence}.
	
	Besides $I^S_1$ and $I^S_2$, we use an additional
	label $\bot$, indicating that no value is available for that pixel due to occlusion. The label set for the MRF is then $\mathcal{L} = \{\bot, 1, 2\}$, where $x_p = 1$ or $x_p = 2$
	indicate that the pixel is copied from
	$I^S_1$ or $I^S_2$, and a label of $x_p = \bot$ indicates that the pixel is occluded by an object in all input images and therefore cannot be accurately reproduced.\footnote{Here we present only the two-image case. The generalization to the multi-image case follows directly and does not change any of the terms; it only increases the label space.}
	
	Given this MRF, we solve for a labeling $x$\ignore{ with the expansion
		move algorithm~\cite{BVZ:PAMI01},} using an objective function that,
	in addition to the traditional data and smoothness terms $E_d$ and
	$E_s$, contains three new terms that we introduce here: a \emph{cropping
		term} $E_c$, a \emph{duplication term} $E_r$, and an \emph{occlusion
		term} $E_o$, which are presented in Section~\ref{subsec:new_terms}, following a
	brief review of the traditional terms. Using a 4-connected adjacency 
	system $\mathcal{N}$ and tradeoff coefficients 
	$\lambda_d, \lambda_s, \lambda_c, \lambda_r, \lambda_o, \delta$, the final 
	energy is then given by:
	
	\begin{equation}
	\begin{aligned}
	E(x) = & \lambda_d \sum_{p \in P} E_d(x_p) + \lambda_s \sum_{p, q \in \mathcal{N}} E_s(x_p, x_q) & + \ & \lambda_c \sum_{\ell \in \calL}\sum_{o \in O_\ell} E_c(x; o, \ell)\ + \\
	& \lambda_r \!\!\!\!\!\!\!\! \sum_{(o_1, o_2) \in \calM(O_1, O_2)} \!\!\!\!\!\!\!\! E_r(x ; o_1, o_2) & + \ & \lambda_o \!\!\!\!\!\!\!\! \sum_{(o_1, o_2) \in \calM(O_1, O_2)} E_o(x ; o_1, o_2) \label{eq:1}
	\end{aligned}
	\end{equation}
	
	\subsubsection{Data term \boldmath$E_d(x_p)$} This term is given by 
	\[
	E_d(x_p) = 
	\begin{cases}
	0, &\qquad x_p \neq \bot \wedge M_{x_p}(p) = 1, \\
	1, &\qquad x_p \neq \bot \wedge M_{x_p}(p) = 0, \\
	1 + \delta, &\qquad x_p = \bot
	\end{cases} 
	\]
	
	This term penalizes choosing a pixel in the output from an input image $i$ if the pixel is not in the mask ($M_i(p) = 0$), or for declaring a pixel occluded. The $\delta$ parameter determines how strongly we
	prefer to leave a pixel empty rather than label it as occluded, and is discussed further in the definition of
	the occlusion term $E_o$ below. There is no preference between the two source images.
	
	\subsubsection{Smoothness term \boldmath$E_s(x_p, x_q)$} To define this term we need the following notation:  $\calC(p, q, r) = \{k \mid \min(\|k - p\|_1, \|k - q\|_1) \le
	r\}$ is the set of pixels within L1 distance $r$ of either pixel $p$ or
	$q$, describing a local patch around adjacent pixel $p$ and $q$, while
	$I_{max} = \max_{p,q}\sum_{k \in \calC(p, q, r)} \|I_{1}^S(k) - I_{2}^S(k)\|$. Writing exclusive-or as $\oplus$, our smoothness term is
	\[
	E_s(x_p, x_q) = 
	\begin{cases}
	0, &\qquad x_p = x_q, \\
	I_{max}, &\qquad x_p = \bot \oplus x_q = \bot, \\
	\sum_{k \in \calC(p, q, r)} \|I^S_{x_p}(k) - I^S_{x_q}(k)\|, &\qquad \text{else}. \\
	\end{cases} 
	\]
	Note that our term for the case
	$x_p = \bot \oplus x_q = \bot$ discourages the MRF
	from transitioning into the occluded label.
	
	In general, $E_s$ penalizes the local photometric difference for a
	seam between pixels $p$ and $q$ when $x_p \ne x_q$. In the special
	case where $r=0$, $\calC(p, q, r) = \{p, q\}$, and the cost of the seam here is
	$\lambda_s (\|I^S_{x_p}(p) - I^S_{x_q}(p)\| + \|I^S_{x_p}(q) -
	I^S_{x_q}(q)\|)$ as in most seam finding algorithms. Values of $r > 0$
	will lead to larger local patches.
	
	\subsection{Our new MRF terms}
	\label{subsec:new_terms}
	
	\subsubsection{Cropping term $E_c$} We introduce a term that penalizes seams that cut through an object 
	$o \in O_\ell$, with cost proportional to the length of the
	seam.\footnote{More precisely, seam means a transition from label
		$\ell$ to non-$\ell$ in particular here, not the transition between
		two arbitrary labels.}
	
	\begin{math}
	E_c(x; o, \ell) = \sum_{p\in o} \sum_{q\in o} [x_p = \ell, x_q \ne \ell]  
	\end{math}
	
	The value of this term is $0$ exactly when object $o$ is either drawn entirely
	from $I^S_\ell$, or not present at all in the final stitching result 
	($x_p = \ell, \forall p \in o$ or $x_p \ne \ell, \forall p \in o$, respectively). As defined, this results in $|o|^2$ pairwise terms, which may cause the optimization to be intractable in practice. As a result, we use an approximation of this term in the experiments, discussed in \ref{sub:optimization}.

	\begin{figure}
		\begin{subfigure}[b]{.24\textwidth}
			\centering
			\includegraphics[height=60px]{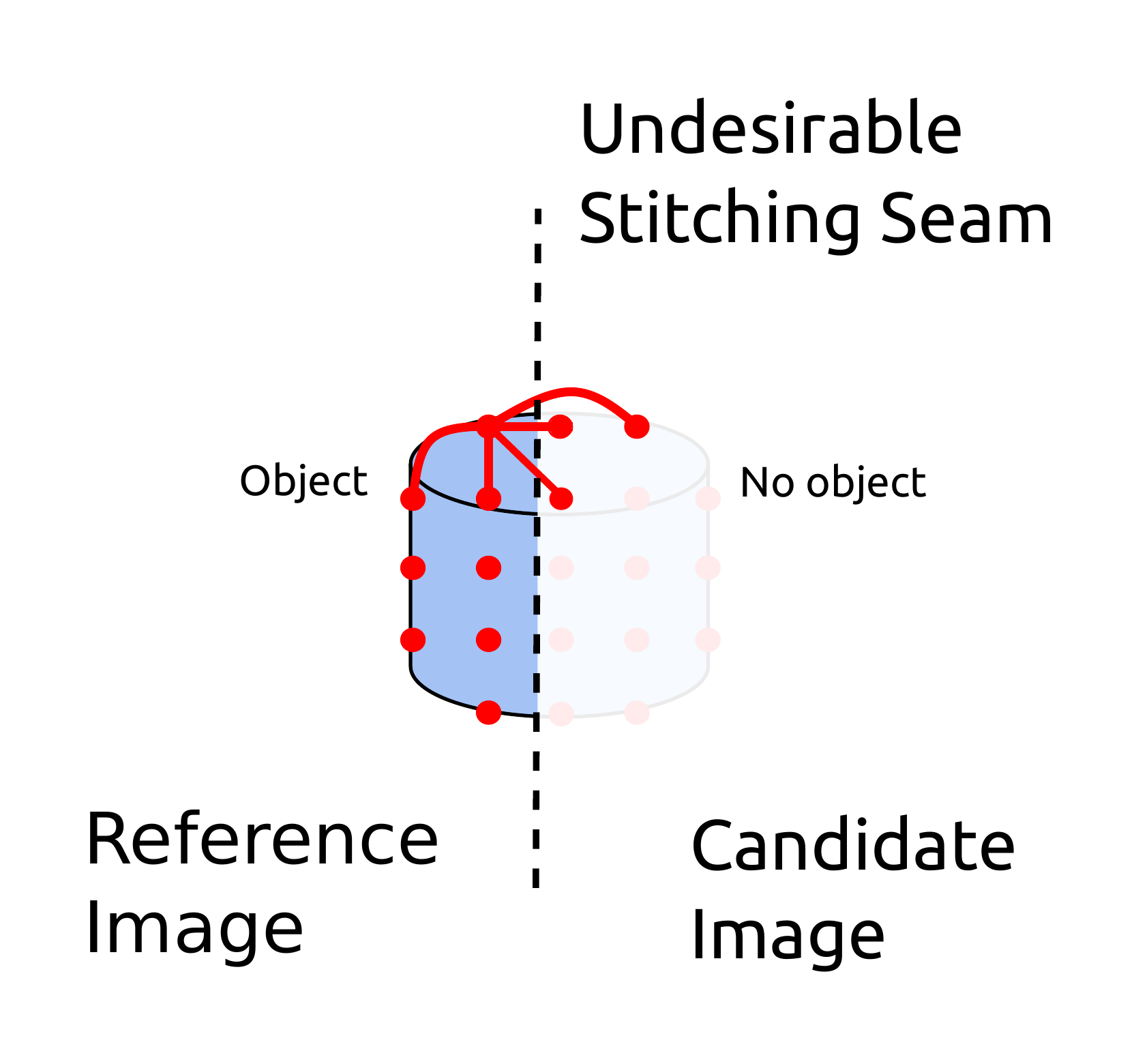}
			\caption{}
			\label{fig:tearf}
		\end{subfigure}
		\begin{subfigure}[b]{.24\textwidth}
			\centering
			\includegraphics[height=54px]{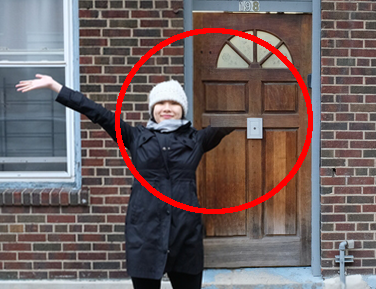}
			\caption{}
			\label{fig:teareg}
		\end{subfigure}
		\begin{subfigure}[b]{.24\textwidth}
			\centering
			\includegraphics[height=58px]{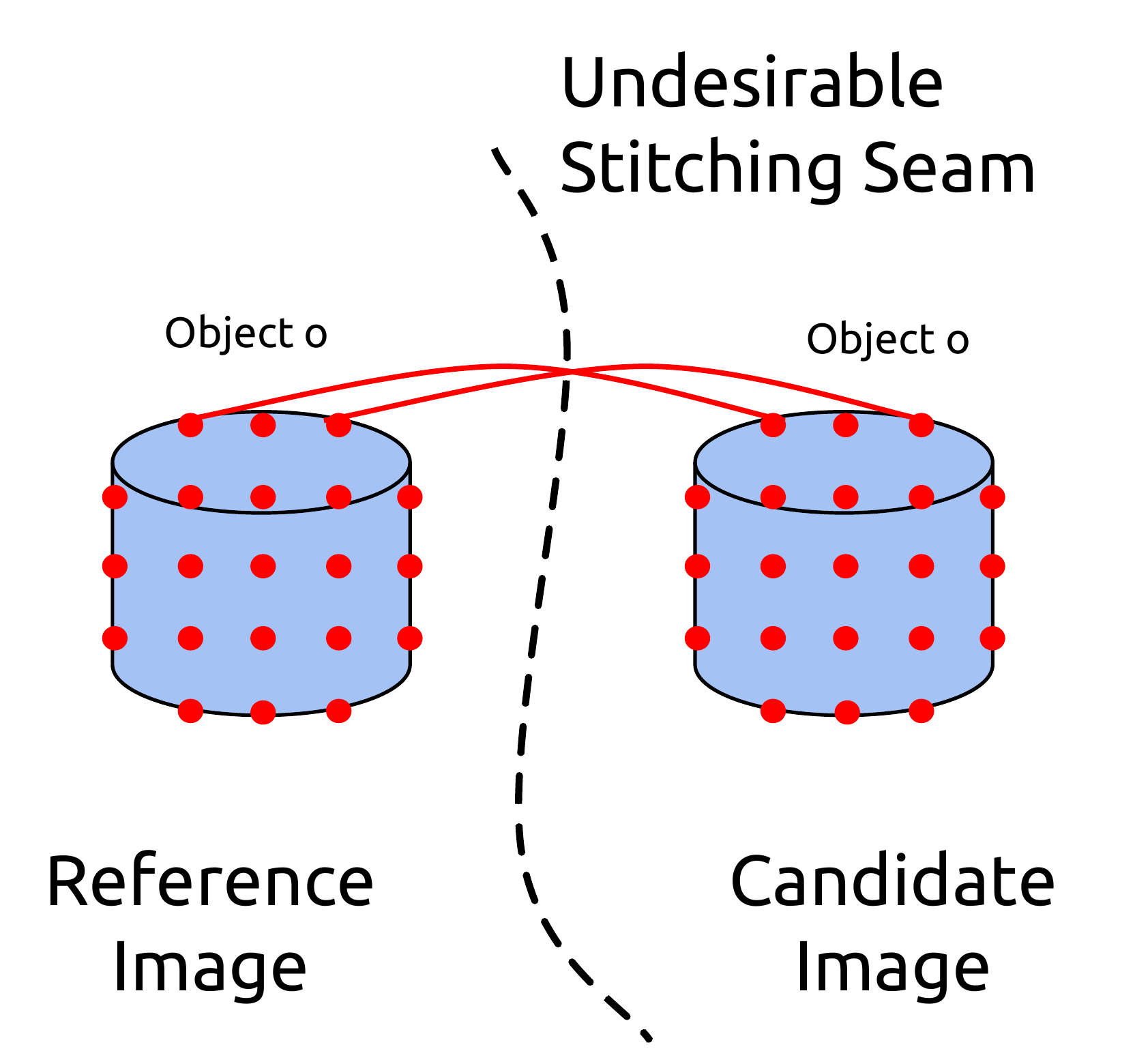}
			\caption{}
			\label{fig:duplf}
		\end{subfigure}
		\begin{subfigure}[b]{.24\textwidth}
			\centering
			\includegraphics[height=48px]{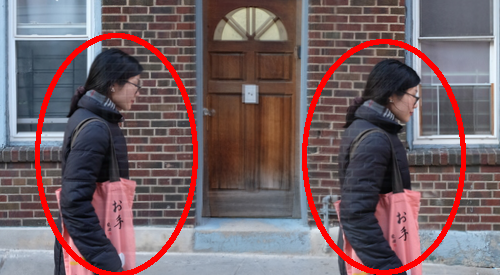}
			\caption{}
			\label{fig:dupleg}
		\end{subfigure}
		\caption{Figure \ref{fig:tearf} depicts our crop term. We use
			pairwise terms to penalize any seam that cuts through an
			object. Figure \ref{fig:teareg} depicts a crop error created
			by Photoshop. Figure \ref{fig:duplf} depicts our duplication
			term. We use pairwise terms to penalize any seam that results in the
			same object appearing in two different locations on the final
			mosaic. Figure \ref{fig:dupleg} depicts a duplication error created
			by NIS.}
	\end{figure}
	
	Note that since this term is a penalty rather than a hard constraint, the tradeoff
	between the smoothness term $E_s$ and this term $E_c$ will still allow us
	to cut through an object if it sufficiently benefits photometric consistency.
	
	%We experimented with a softer constraint on the tear term that
	%related the tear penalty to the difference between $I_1$ and $I_2$ in
	%a large patch.  Since we are trying to manipulate an object, we need
	%to maintain a certain structure and must therefore make the seam
	%comparison on a much larger basis - one that takes into account the
	%structure of the object.
	
	%The primary way to do this was different values for $\lambda_t$. One
	%value was the $SSD\_of\_RGB$ of all $p\in o$ for labels $0$ and
	%$1$. We tested also tested local patches centered around the pixels
	%in the pairwise term.
	
	%TODO(chi): WRITE EQUATIONS HERE!
	
	%However, in the end we found that setting $\lambda_t$ to a large
	% constant performed well and avoided tuning.\ read, we ran out of
	% time. - though I could probably make these work given a half a day
	% or something
	
	\subsubsection{Duplication term $E_r$} Our term discourages duplication when $o_1$ in $I^S_1$ and
	$o_2$ in $I^S_2$ are known to refer to the same object, and is defined as
	
	\begin{math}
	E_r(x; o_1, o_2) = \sum_{(p, q) \in m(o_1, o_2)} [x_p = 1 \land x_q = 2].
	\end{math}
	
	\noindent Here $m(o_1, o_2) \in o_1 \times o_2$ are the pixel-level
	correspondences between objects $o_1$ and $o_2$. $(p, q) \in m(o_1, o_2)$ 
	represent the same point in the real world, so the final stitching result
	should not include both pixel $p$ from $o_1$ and pixel $q$ from $o_2$. Note that this term includes a potentially complicated function $m$ that calculates dense pixel correspondences; as a result, we use an approximation of this term in the experiments, discussed in \ref{sub:optimization}. 
	
	\subsubsection{Occlusion term $E_o$} This term promotes the occlusion label by penalizing the use of out-of-mask
	labels in areas of the image where duplicate objects were detected:
	
	% allows the occlusion label to
	% occur only if a duplicated object covers an area in the input image that 
	% has no other labels with pixel values:
	
	% PREVIOUS VERSION
	
	% \begin{align}
	% \begin{split}
	% E_o(x; o_1, o_2) = &(\lambda_d+\delta) \sum_{p\in P} [x_p = \bot] + 2\delta \sum_{\ell\in \{1, 2\}} \sum_{o\in % O_\ell} \sum_{p\in o} [M_\ell(p) = 0 \land x_p = \ell] 
	% \end{split}
	% \end{align}
	
	% END PREVIOUS VERSION
	
	\begin{align}
	\begin{split}
	E_o(x; o_1, o_2) = 2\delta \sum_{\ell\in \{1, 2\}} \sum_{p\in o_{\ell}} [M_\ell(p) = 0 \land x_p = \ell] 
	\end{split}
	\end{align}
	
	% where $\lambda_d$ and $\lambda_s$ are the weighting terms for the data term $E_d$ and the smoothing term $E_s$
	% described above, and $\delta$ is a parameter of our method.
	
	\noindent where $\delta$ is the same parameter used to penalize the selection of the $\bot$ label in $E_d$.
	For the intuition behind this term, consider the case where $o_1$ and 
	$o_2$ are corresponding objects in $I^S_1$ and $I^S_2$, and $M_2(p) = 0$ for $p \in o_1$. Then we
	must either select label $1$ for the pixels in $o_1$ or 
	declare the pixels occluded. The data term $E_d$ ensures that the occlusion label will
	normally give higher energy than a label which is out of mask. However, in the presence of a duplicated object,
	the occlusion term $E_o$ increases the energy of the out of mask term since $2\delta > \delta$, resulting in the occlusion label being
	selected instead. Note, we typically set $\lambda_o = \lambda_d$.
	
	%Lowering the energy for the occlusion label would also work with 2
	% input images but does not generalize. The approach we do out of mask
	% b/c we don't want to set \lambda_d too high and if we decrease and
	% have a low-ish \lambda_d, then pairwise makes more of a difference?
	% need to think about this.
	
	\subsubsection{Generalization to 3 or more images.} With multiple inputs, one
	image acts as the reference and the others become candidates. We then
	calculate registrations in the same manner as before, then pass to the seam
	finding phase the reference image and the candidate registrations:
	$I_1$ and $\warp_2(I_2),\ldots,\warp_n(I_n)$. We calculate correspondence
	for all pairs of images. When establishing correspondence between
	objects, we make sure that correspondence acts as an equivalence
	relation. The primary difference between the two and three input image
	case is transitivity. If three objects violate transitivity, we increase the correspondence threshold until the property
	holds. While other schemes could be imagined to ensure consistency,
	experimentally, we have yet to see this be violated.
	
	\subsection{Optimization}
	\label{sub:optimization}
	
	% \subsubsection{Local Energy terms.}
	The cropping term $E_c$ above has dense connections between each pixel $p \in I^S_1$ and $q \in I^S_2$, which can lead to computational difficulties. Here we introduce a \emph{local energy term} $E_{lc}$ that has fewer connections and is therefore simpler to compute, while experimentally maintaining the properties of the terms introduced above:
	
	\begin{align*}
	E_{lc}(x; o, \ell) = \sum_{p\in o} \sum_{q\in N_p} [x_p = \ell, x_q \ne \ell]  
	\end{align*}
	
	\noindent where $N_p$ is set of neighbors for $p$.
	
	Similarly, the duplication term reported above has a complicated structure based on the matching function over  detected objects. We define the \emph{local duplication term} $E_{lr}$ in terms of $m_b(o_1, o_2)$, which in contrast to $m(o_1, o_2)$, returns the corresponding points of the two \emph{bounding boxes} around objects $o_1$ and $o_2$, where each $p \in o_1$ is bilinearly interpolated to its position in $o_2$ using the corners of the bounding box.
	
	To solve this MRF, we use alpha-expansion \cite{BVZ:PAMI01} with QPBO \cite{QPBO} for the induced binary subproblems. QPBO has been reported to perform well on a variety of computer vision tasks in practice, even when the induced binary subproblem is supermodular \cite{QPBO}.
	
	\subsection{Establishing correspondence between objects}
	\label{subsec:establishing-correspondence}
	
	Our strategy is to consider pairs of objects $o_1, o_2$
	detected in images $I_1$, $\warp(I_2)$ respectively and to compute a
	metric that represents our degree of confidence in their corresponding
	to the same object. We compute this metric for all object pairs over all images and
	declare the best-scoring potential correspondence to be a match if it exceeds a specified threshold. In addition to the \emph{correspondence density} metric used in
	the experiments reported here, we considered and tested several
	different metrics that we also summarize below. In all cases, the category returned by the object detector
	was used to filter the set of potential matches to
	only those in the same category.
	
	\noindent\textbf{Feature point matching.} We tried running SIFT~\cite{Lowe:SIFT99} and DeepMatch~\cite{Weinzaepfel_2013_ICCV}
	directly on the objects identified. 
	These methods gave a large number of correspondences without spatial coherence;
	for example, comparing a car and bike would
	result in a reasonable number of matches but the points in image $I_1$ would match to points very far away in $I_2$. 
	We tried to produce a metric that captured this
	by comparing vector difference between feature points $p$ and $q$
	from $I_1$ to their correspondences $p'$ and $q'$ from $I_2$.
	
	%$\sum_{(p,p'), (q,q')\in C_{1, 2}} |V_{p,q} - V_{p', q'}|$
	
	\noindent\textbf{Correspondence density.} We ran DeepMatch~\cite{Weinzaepfel_2013_ICCV} on the two input
	images and counted the matches belonging to the two
	objects being considered as a match. This number was then divided by
	the area of the first image. Since DeepMatch feature points are roughly uniformly distributed in the first input image, the density of points in the
	area of the first input has an upper bound and it is possible to pick a density threshold that is able to distinguish between matching and non-matching objects, regardless of their size. This is the technique used in the experimental section below.
	
	% and can be much better
	% predicted. In the end, we set a simple threshold on the density of
	% correspondences. Experimentally, we found that there was a large
	% numerical difference between matching objects and non-matching
	% objects. 
	% check the DeepMatch thing?

	\section{Object-centered evaluation of stitching algorithms}
	\label{sec:eval-stitch}
	
	We now discuss the use of object detectors for formalized evaluation
	of stitching algorithms. In general, we assume access to the input
	images and the final output. The available object detectors are run on
	both, and their output used to identify crops, duplication, or
	omissions introduced by the stitching algorithm. The goal of this
	evaluation technique is not to quantify pixel-level discontinuities,
	e.g. slight errors in registration or seam-finding, but rather to
	determine whether the high-level features of the scene, as indicated
	by the presence and general integrity of the objects, are preserved.
	
	In the following, $F$ denotes the final output panorama, $I$ the set
	of input images, and $N_X$ the number of objects detected in an image
	$X$. $N_F$, for instance, would denote the number of objects found by
	a detector in the stitch result. Note that the techniques we
	propose can also be applied in parallel for specific categories of
	objects: instead of a general $O$ and $N_F$, we might consider $O^c$
	and $N_F^c$ for a particular category of objects $c$,
	e.g. humans or cats. Separating the consideration of objects in this way 
	makes object analysis more granular and more likely to identify
	problems with the scene.
	
	\setlength{\fboxrule}{1px}
	\setlength{\fboxsep}{1px}
	\definecolor{orange}{RGB}{255,165,0}
	\definecolor{darkorange}{RGB}{255,110,10}
	\definecolor{salmon}{RGB}{255,160,122}
	\definecolor{orangered}{RGB}{255,69,0}
	\definecolor{pink}{RGB}{255,105,180}
	\definecolor{magenta}{RGB}{255,0,255}
	
	\ignore{
		\begin{figure}
			\begin{subfigure}{1\textwidth}
				\fcolorbox{blue}{white}{
					\includegraphics[height=47px]{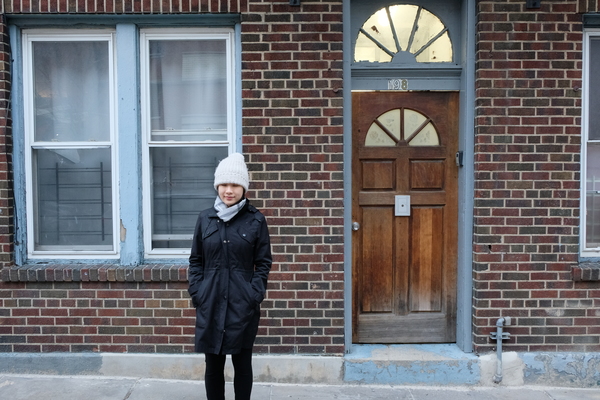}
					\includegraphics[height=47px]{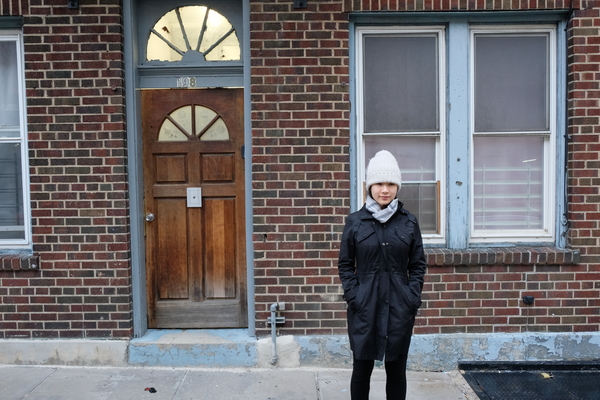}
				}
				\includegraphics[height=47px]{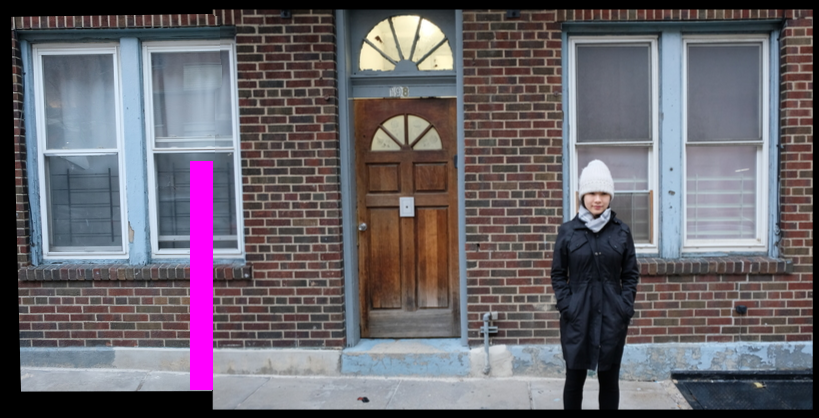}
				\includegraphics[height=47px]{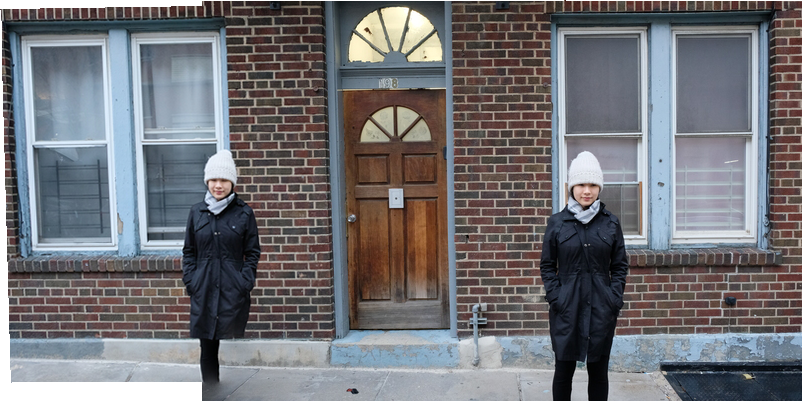}
			\end{subfigure}
			\begin{subfigure}{1\textwidth}
				\fcolorbox{blue}{white}{
					\includegraphics[height=47px]{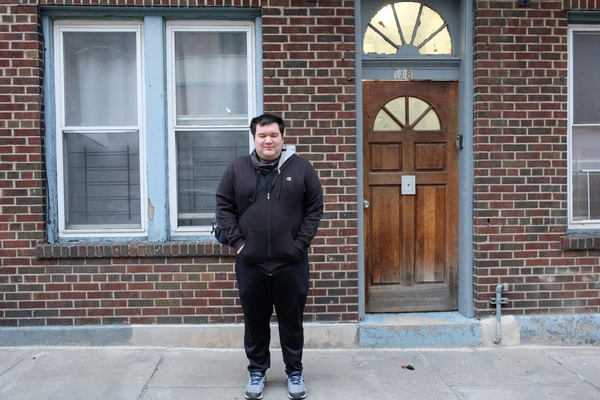}
					\includegraphics[height=47px]{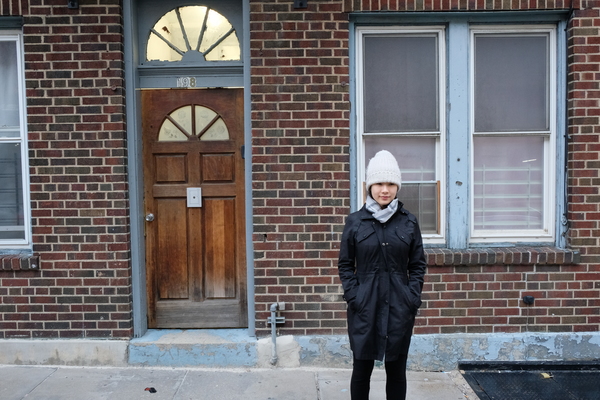}
				}
				\includegraphics[height=47px]{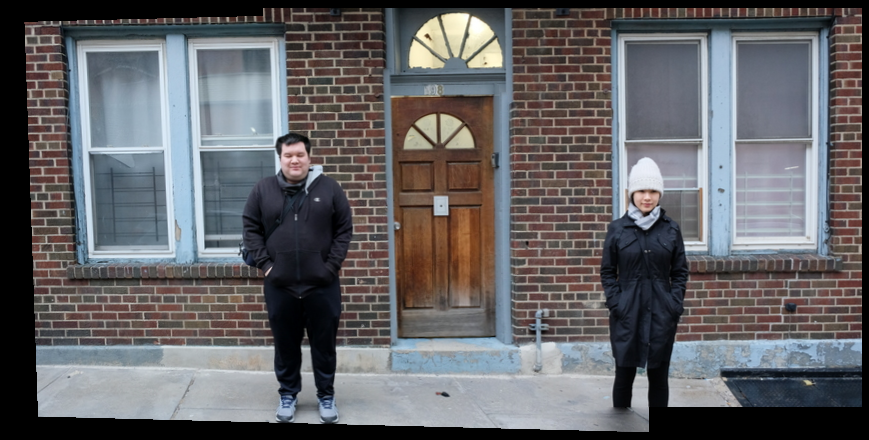}
				\includegraphics[height=47px]{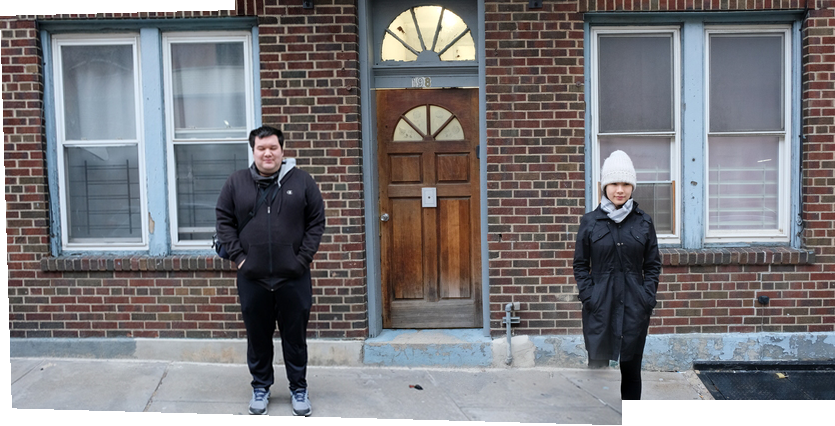}
			\end{subfigure}
			\caption{\label{skiing-evaluation} Object bounding boxes for
				humans detected in given source. Both rows from left to
				right: candidate and reference input images, Our result,
				Photoshop result. Note that the magenta in top row is
				detected occlusion. The density term for the two objects in
				the top row is 0.630826 and for the two objects in the
				bottom row is 0.0517842.}
		\end{figure}
	}

	\setlength{\fboxsep}{1px}

	\subsection{Penalizing omission and duplication}
	
	We first attempt to evaluate the quality of a stitch through the number of objects $N$ detected in the input
	images and the final output. We generalize $\calM(O_1, \ldots, O_n)$ to apply to an
	arbitrary number of input images, denoting corresponding object
	detections across a set of images $I_1, \ldots, I_n$. The techniques
	discussed above for establishing correspondences between objects can
	easily be generalized to multiple images and used to formulate an
	expression for the expected number of objects in the final stitch
	result. In particular, the expected object count for a hypothetical
	ideal output image $F^*$ is given by the number of ``equivalence
	classes'' of objects found in the input images for the correspondence
	function under consideration: all detected objects are expected to be
	represented at least once, and corresponding objects are expected to
	be represented with a \emph{single} instance.
	
	% \begin{align}
	% begin{split}
	% N_{F^*} = \sum_{I_i\in I} |O_i| &-  \sum_{I_{i_1}\neq I_{i_2}} |\calM(O_{i_1}, O_{i_2})|  +  \sum_{I_{i_1}\neq I_{i_2}\neq I_{i_3}} |\calM(O_{i_1}, O_{i_2}, O_{i_3})| \\
	%  &- \qquad\qquad \ldots \qquad\qquad\hspace{-.3em} +  \sum_{I_1\neq\ldots\neq I_n} |\calM(O_{i_1}, \ldots, O_{i_n})|\
	% \end{split}
	% \end{align}
	
\begin{figure}
		\begin{subfigure}{.33\textwidth}
			\centering
			\includegraphics[height=100px,width=110px]{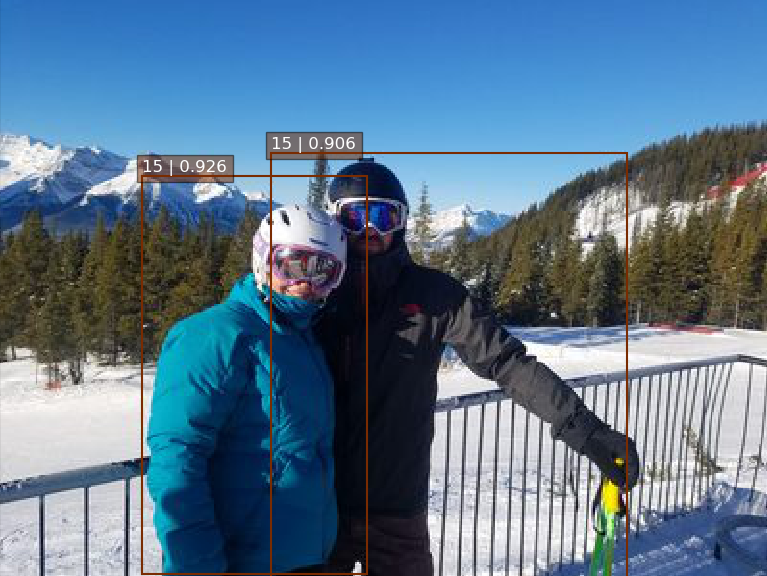}
			\caption{Input image}
		\end{subfigure}
		\begin{subfigure}{.33\textwidth}
			\centering
			\includegraphics[height=100px,width=110px]{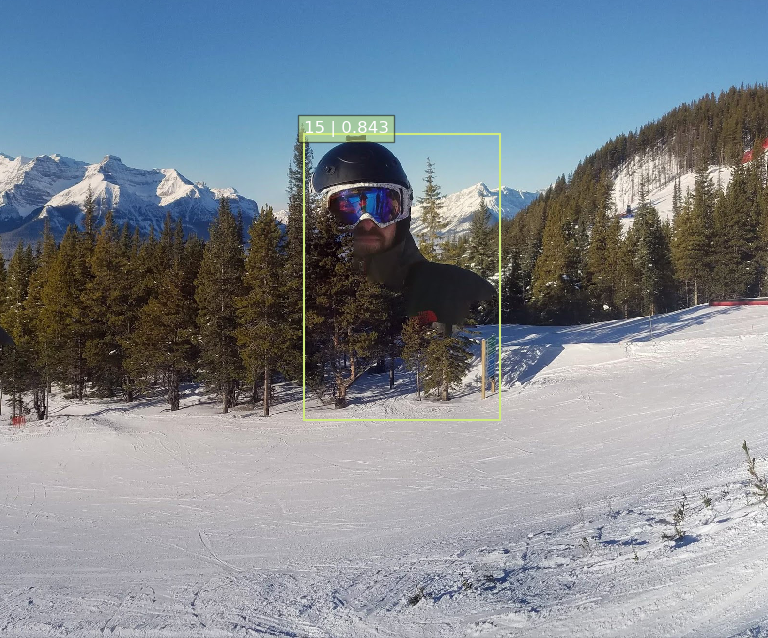}
			\caption{Google Photos (0.1140)}
		\end{subfigure}
		\begin{subfigure}{.33\textwidth}
			\centering
			\includegraphics[height=100px,width=110px]{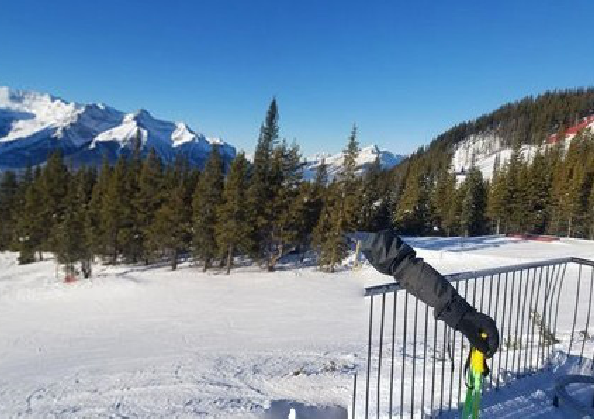}
			\caption{Photoshop (no object)}
		\end{subfigure}
		
		\begin{subfigure}{.5\textwidth}
			\centering
			\includegraphics[height=100px]{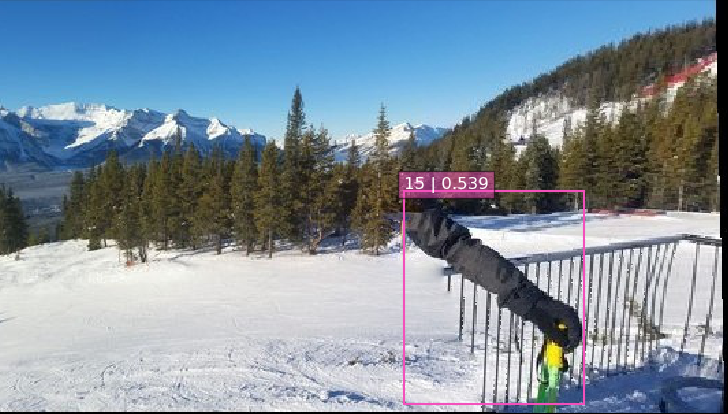}
			\caption{APAP (0.1203)}
		\end{subfigure}
		\begin{subfigure}{.5\textwidth}
			\centering
			\includegraphics[height=100px]{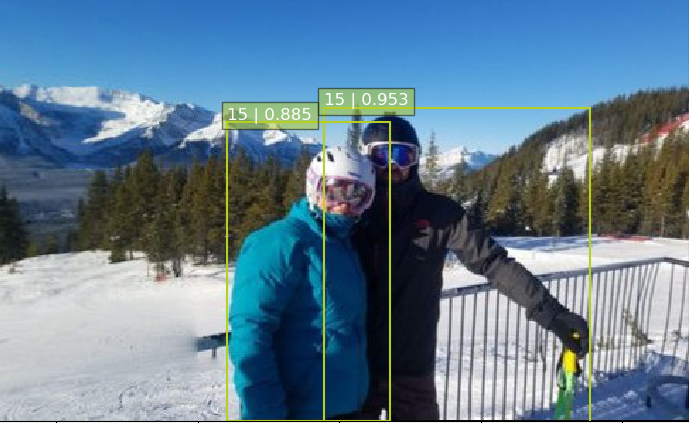}
			\caption{Our result (left 0.4882, right 0.93380)}
		\end{subfigure}
		\caption{\label{fig:ski_eval} Visualizations for object
			bounding boxes for humans detected in given source. Final mosaics have been altered for
			space reasons, but no bounding boxes were removed. The MS-SSIM are listed in parenthesis after the method name.  $N_F - N_{F^*}$ is as follows (b) -1, (c) -2, (d) -1, and (e) 0. }
	\end{figure}

	For a good stitching output $F$, we expect $N_{F} = N_{F^*}$. Note that $N_{F}
	> N_{F^*}$ or $N_{F} < N_{F^*}$ imply omissions or duplications,
	respectively. In Figure \ref{fig:ski_eval}, a human detector finds objects 
	in only one image and $\calM(O_1, O_2) = \emptyset$; therefore, we have 
	that $N_{F^*} = 2$ for the category of humans. When run on the output of 
	Photoshop or APAP, however, only one human is found, 
	giving $N_{F} < N_{F'}$ and indicating an omission.
	
	Other approaches exist for detecting omission or duplication that do
	not require computing the potentially complicated $\calM$
	function. For example, it can be inferred that an object has been
	omitted in the output if it contains fewer objects than any of
	the inputs: $N_F < \max_{I_i\in I}(|O_i|)$. Similarly, a duplication 
	has occurred if more objects are identified in the output
	than the total number of objects detected in all of the input images:
	$N_F > \sum_{I_i\in I} N_{I_i}$. While this may seem to be a weak
	form of inference, it proves sufficient in Figure \ref{fig:ski_eval}:
	the maximum number of humans in an input image is $2$, but only one is found in the Photoshop and APAP results, indicating an omission.
	
	Unfortunately, while duplication almost always indicates an error
	in an output $F$, the situation is not as clear-cut with
	omissions. Objects that are not central to the scene or that are not
	considered important by humans for whatever reason can often be
	omitted without any negative effect on the final mosaic.
	
	\subsection{Cropping}

	\setlength{\fboxsep}{-2px}
	
	\begin{figure}
		\begin{subfigure}{.42\textwidth}
			\centering
			%		\fcolorbox{blue}{white}{
			\includegraphics[height=46px]{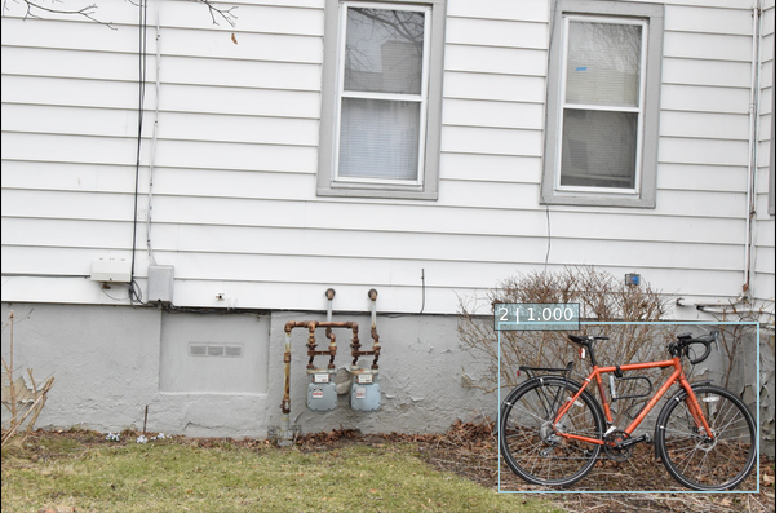}
			\includegraphics[height=46px]{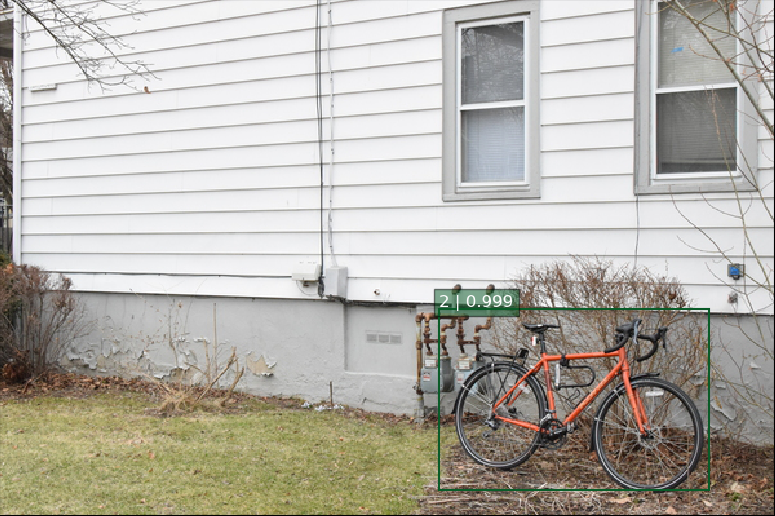}
			\caption{Input images}
			%		}
		\end{subfigure}%
		\begin{subfigure}{0.2\textwidth}
			\centering
			\includegraphics[height=49px]{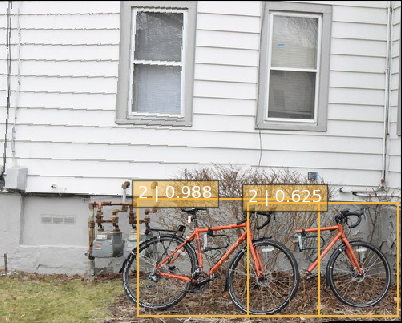}
			\caption{APAP}
		\end{subfigure}
		\begin{subfigure}{0.2\textwidth}
			\centering
			\includegraphics[height=49px]{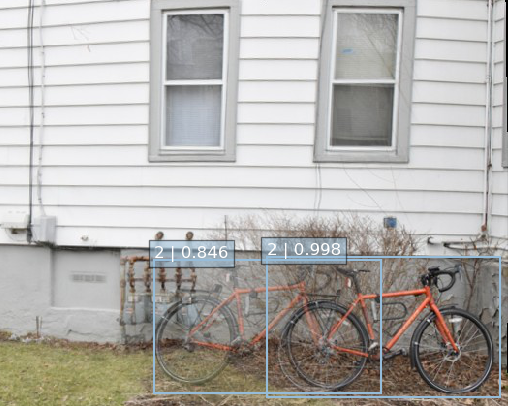}
			\caption{NIS}
		\end{subfigure}%
		\begin{subfigure}{0.2\textwidth}
			\centering
			\includegraphics[height=49px]{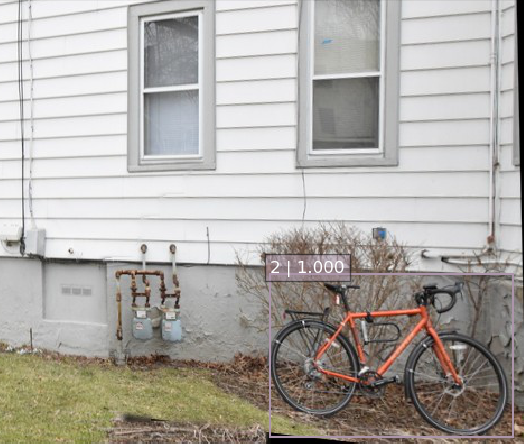}
			\caption{Our result}
		\end{subfigure}
		\caption{Visualizations for object bounding boxes for bikes
			detected in given source.  Final mosaics have been altered for space reasons, but no bounding boxes were removed. Other techniques failed to produce a stitch. The MS\_SSIM are as follows: APAP left (0.1608), APAP right (0.1523), NIS left (0.3971), NIS right (0.1771), Ours (\textbf{0.8965}).  $N_F - N_{F^*}$ is as follows (b) 1, (c) 1, and (d) 0 }
	\end{figure}

	\setlength{\fboxrule}{1pt}
	\setlength{\fboxsep}{2pt}
	\begin{figure}
		\begin{subfigure}{.33\textwidth}
			\centering
			\includegraphics[height=90px,width=100px]{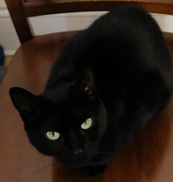}
			\caption{Input image}
		\end{subfigure}
		\begin{subfigure}{.33\textwidth}
			\centering
			\includegraphics[height=90px,width=100px]{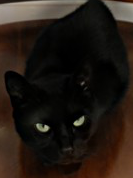}
			\caption{Autostitch (0.4575)}
		\end{subfigure}
		\begin{subfigure}{.33\textwidth}
			\centering
			\includegraphics[height=90px,width=100px]{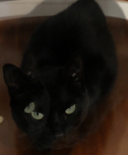}
			\caption{NIS (0.5201)}
		\end{subfigure}
		\begin{subfigure}{.33\textwidth}
			\centering
			\includegraphics[height=90px,width=100px]{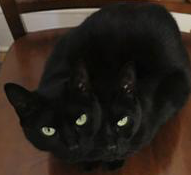}
			\caption{Photoshop (0.5099)}
		\end{subfigure}
		\begin{subfigure}{.33\textwidth}
			\centering
			\includegraphics[height=90px,width=100px]{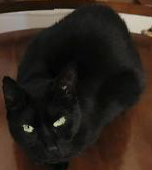}
			\caption{APAP (0.4773)}
		\end{subfigure}
		\begin{subfigure}{.33\textwidth}
			\centering
			\includegraphics[height=90px,width=100px]{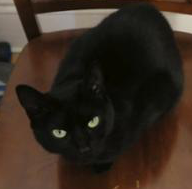}
			\caption{Our result (\textbf{0.6769})}
		\end{subfigure}
		\caption{\small Object bounding boxes for cats detected in given source. MS\_SSIM is included in parenthesis. Autostitch applies a warp that alters the cat's facial shape. NIS contains ghosting. Photoshop duplicates part of the cat's face. APAP applies a warp that alters the cat's facial shape.}
	\end{figure}
	
	Object detectors can be used to detect crops in two ways: major crops,
	which make the object unrecognizable to the detector, are interpreted
	by our system as omissions, and detected as described above. Even
	objects that are identifiable in the final output, however, may be
	partially removed by a seam that cuts through them or undergo
	unnatural warping. A different approach is therefore needed in order
	to detect and penalize these cases. Here we consider two options that
	differ in whether they consider the results of the object detector on
	the input images: the first directly compares the detected objects
	from the inputs and output, while the second is less sensitive to the
	choice of object detector and instead uses more generic template
	matching methods.
	
	For both of the methods, we note that some warping of the input image
	is usually necessary in order to obtain a good alignment, so image
	comparison techniques applied to the original input image and the
	output image are unlikely to be informative. However, given an input
	image $I_i$ and the output $F$ it is possible to retrospectively
	compute a set of plausible warps $\warp(I_i)$ and apply image
	comparison operators to these. Our approach therefore does not require
	access to the actual warp used to construct the stitch, but it
	can of course be used to increase the accuracy of our methods if it is
	available. 
	
	% Emil: Unclear what this means - to the supplementary materials? to another paper?
	% Computation of plausible warps is deferred.

	\noindent\textbf{Cropping detection with direct object comparison.} This approach
	implicitly trusts the object detector to give precise results on both
	the input images and the output image. The object detector is run for
	$F$ and for all of the plausible registration candidates that have
	been determined for the various $I_i$. We then run Multiscale Structural Similarity (MS-SSIM) \cite{MSSIM} for all
	of the correspondences among the detected objects (determined as
	discussed in Section~\ref{subsec:establishing-correspondence}), and
	use the average and maximum values of these metrics as our final
	result. Any reasonable image similarity metric can be used in this
	approach, including e.g. deep learning techniques.
	
	\noindent\textbf{Cropping detection with template matching.} This
	metric is less sensitive to the choice of object detector. Instead of
	applying it to all of the warped input images, we apply it only to the
	result image. The region of the output where the object is detected is
	then treated as a template, and traditional template matching
	approaches are used to compare the object to the reference image $I_1$
	and any plausible registrations.
	
	We have experimented with these metrics to confirm that these values
	match our intuitions about the handling of objects in the stitch
	result. We provide some examples and their evaluation values (maximum MS-SSIM with direct object comparison) in the captions of the figures above.
	
	\section{Experimental results for stitching}
	\label{sec:data}
	
	Our goal is to stitch difficult image sets that give rise to noticeable errors with existing approaches. Unfortunately, there is no standard data set of
	challenging stitching problems, nor any generally accepted metric to
	use other than subjective visual evaluation. We therefore follow the
	experimental setup of~\cite{Zhang_2014_CVPR}, who both
	introduce a stitching technique that is able to stitch a difficult class of
	images, and also present a set of images that cause
	previous methods to introduce duplications and cropping. For competitors we consider Photoshop 2018's ``Photomerge''
	stitcher, APAP \cite{zaragoza2013projective}, Autostitch
	\cite{Brown_IJCV_2007}, and NIS\cite{Chen2016}. Following the
	approach of~\cite{Zhang_2014_CVPR}, we extend APAP with a seam finder.
	
	\noindent\textbf{Experimental Setup.} We tried several methods for feature
	extraction and matching, and found that
	DeepMatch~\cite{Weinzaepfel_2013_ICCV} gave the best results. It was
	used in all examples shown here.  The associated DeepFlow solver was
	used to generate flows for the optical flow-based warping. The QP
	problems used to obtain the mesh parameters and determine candidate
	warpings $\omega_i$ were solved with the Ceres solver
	\cite{ceres-solver}. For object detection, we experimented with the
	Mask R-CNN \cite{Ren:RCNN:PAMI17}
	and SSD
	\cite{Liu:SSD16} systems. Both were found to give good
	performance for different types of objects.
	
	\noindent\textbf{Ablation Study.} We performed an ablation study on the pairwise terms in the seam finding stage and found that all terms are necessary and perform as expected. These results are available with the rest of the data as indicated below.
	
	In the remainder of this section, we review several images from our test
	set and highlight the strengths and weaknesses of our technique, as well
	as those of various methods from the literature. All results shown use the same parameter set. Data, images and additional material omitted here due to lack of space are available online.\footnote{See \url{https://sites.google.com/view/oois-eccv18}.}
	
	%% \noindent \textbf{Cat sitting.}
	%% (Figure~\ref{fig:data_results/2_images/cat_orin_sitting}) A \textit{cat}
	%% sitting on a chair taken from two different perspectives. 
	
	%% \noindent \textbf{Pet store.}
	%% (Figure~\ref{fig:data_results/2_images/pet_store_pet_store}) A \textit{person}
	%% in front of a display stand moves from one shot to the next.
	
	%% \noindent \textbf{Bike.}
	%% (Figure~\ref{fig:data_results/2_images/bike_tear}) A \textit{bicycle}
	%% leaning against the side of a house, shot from two different
	%% perspectives.
	
	%% \noindent \textbf{Shiba bag walk.}
	%% (Figure~\ref{fig:data_results/2_images/shiba_bag_walk}) A
	%% \textit{person} walking next to a house, shot at two different
	%% locations.
	
	%% \noindent \textbf{Bottles.}
	%% (Figure~\ref{fig:data_results/2_images/bottles_bottles12}) Two
	%% \textit{bottles} and various other objects on a table, shot from two
	%% different perspectives.
	
	%% \noindent \textbf{Three images.} (Figure~\ref{fig:3-image}) Three
	%% images of a \emph{person} standing in different locations in front of a
	%% building.
	
	\newcommand{\resultsfigure}[5]{
		\begin{center}
			\begin{minipage}{0.31\textwidth}
				\setlength{\abovecaptionskip}{1pt}
				\setlength{\belowcaptionskip}{1pt}
				\includegraphics[width=0.48\textwidth]{#1/input_can.jpg}
				\includegraphics[width=0.48\textwidth]{#1/input_ref.jpg}
				\captionof*{figure}{(a) Inputs}
			\end{minipage}
			\begin{minipage}{0.31\textwidth}
				\includegraphics[width=\textwidth]{#1/#4}
				\captionof*{figure}{(b) #5}
			\end{minipage}
			\begin{minipage}{0.31\textwidth}
				\includegraphics[width=\textwidth]{#1/#2.png}
				\captionof*{figure}{(c) Our result}
			\end{minipage}
			\captionof{figure}{\label{fig:#1}#3}
		\end{center}
	}
	
	% move this up
	%\resultsfigure{data_results/2_images/ski_left}{ours_blend}{``Ski Image'' dataset. From reddit.com.}{photomerge.png}{Photomerge result}
	\ignore{
		\resultsfigure{data_results/2_images/cat_orin_sitting}{ours_blend}{``Cat Sitting'' dataset. NIS produces substantial ghosting on the foreground objects. Our result is plausible.}{nis.png}{NIS result}
		% RDZ: cut the below
		\resultsfigure{data_results/2_images/shiba_bag_walk}{ours_blend_alpha}{``Shiba Bag Walk'' dataset. Autostitch duplicates the foreground person and has ghosting on the door. Our result is plausible.}{autostitch.jpg}{Autostitch result}
		
	}

	\begin{center}
		\begin{minipage}{0.38\textwidth}
			\centering
			\setlength{\abovecaptionskip}{1pt}
			\setlength{\belowcaptionskip}{1pt}
			\includegraphics[width=0.7\textwidth,angle=90,height=70px,origin=c]{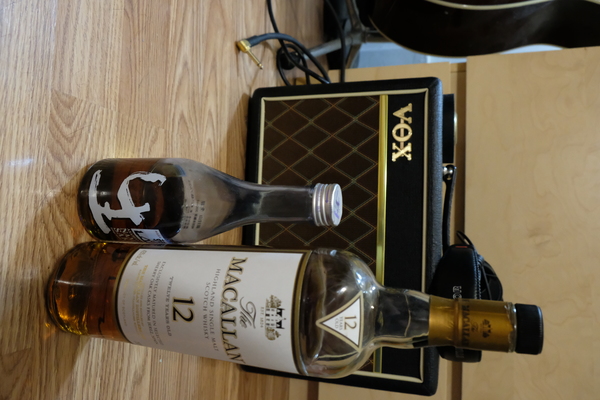}
			\includegraphics[width=0.7\textwidth,angle=90,height=70px,origin=c]{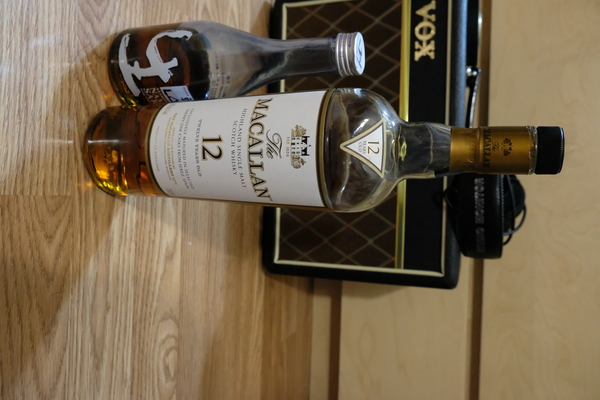}
			\captionof*{figure}{(a) Inputs}
		\end{minipage}
		\begin{minipage}{0.3\textwidth}
			\centering
			\includegraphics[width=\textwidth,height=90px]{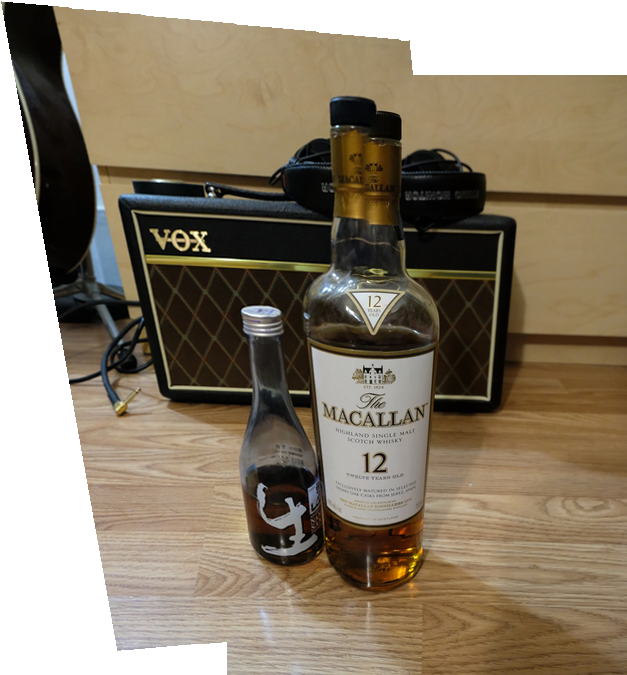}
			\captionof*{figure}{(b) Photoshop result}
		\end{minipage}
		\begin{minipage}{0.3\textwidth}
			\centering
			\includegraphics[width=\textwidth,height=90px]{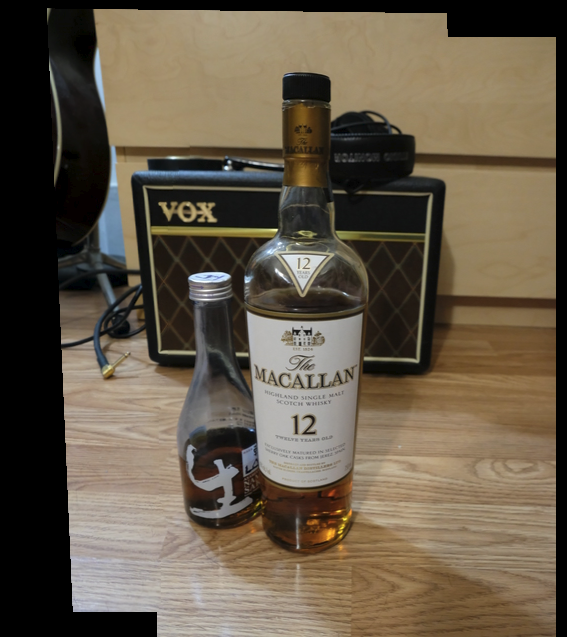}
			\captionof*{figure}{(c) Our result}
		\end{minipage}
		\captionof{figure}{\label{fig:bottles12}``Bottle'' dataset. Photoshop duplicates the neck of the bottle and the headphones. Our result is plausible.}
	\end{center}

	\resultsfigure{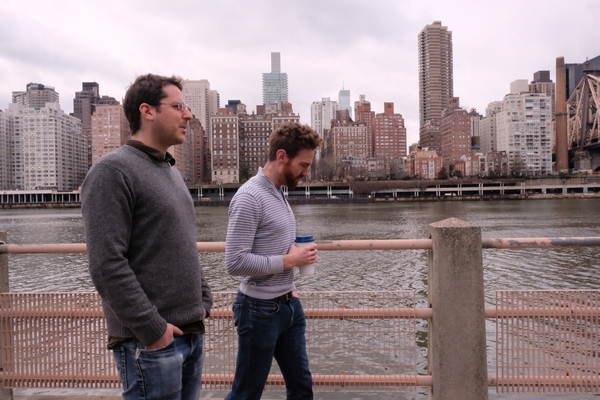}{ours_blend}{``Walking'' dataset.}{photomerge.png}{Photoshop result}
	\resultsfigure{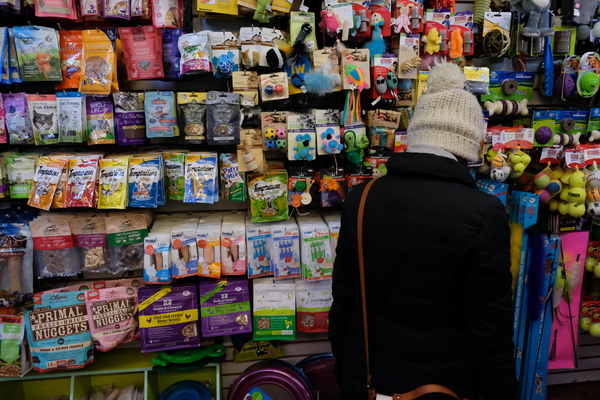}{ours_blend}{``Pet Store'' dataset. 
		Photoshop omits the left arm. Our result is plausible.}{photomerge.png}{Photoshop result}
	\begin{figure}[!ht]
		\begin{minipage}{1\textwidth}
			\begin{subfigure}{1\textwidth}
				\begin{centering}
					\maybeimage{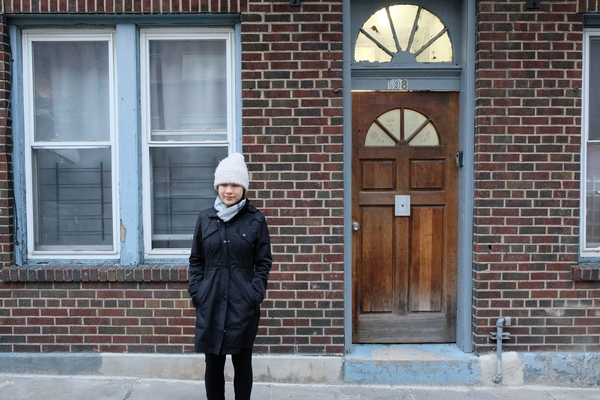}{0.3}
					\maybeimage{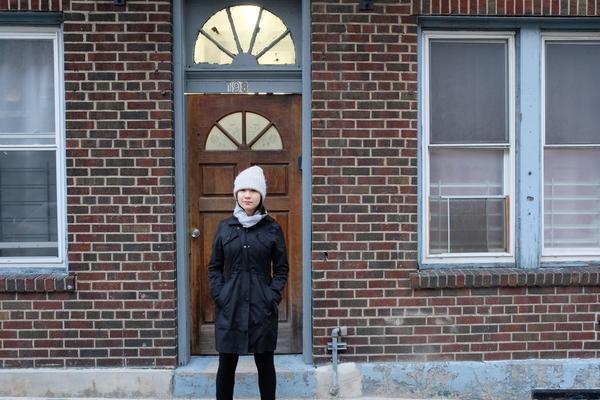}{0.3}
					\maybeimage{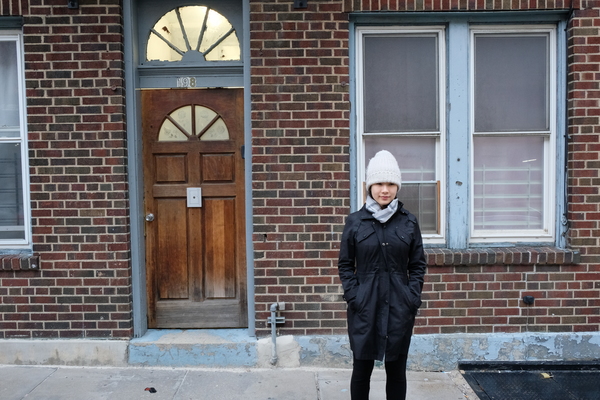}{0.3}
					\caption{Candidate and reference images}
				\end{centering}
			\end{subfigure}
			\begin{subfigure}[t]{.5\textwidth}
				\centering
				\maybeimage{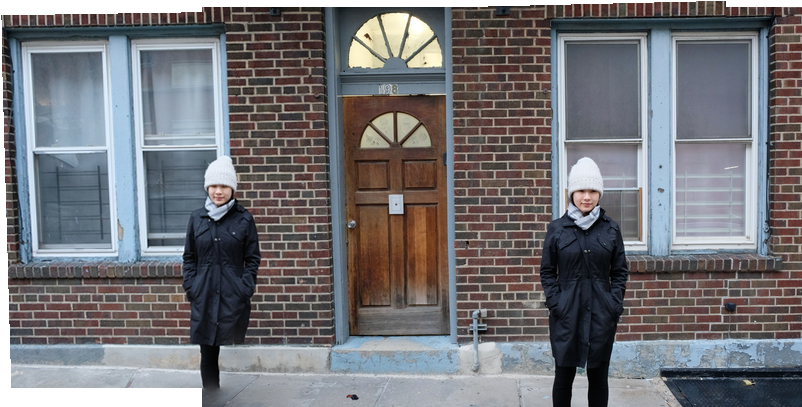}{1.0}
				\caption{\label{fig:3_image_autostitch} Photoshop result}
			\end{subfigure}
			\begin{subfigure}[t]{.5\textwidth}
				\centering
				\maybeimage{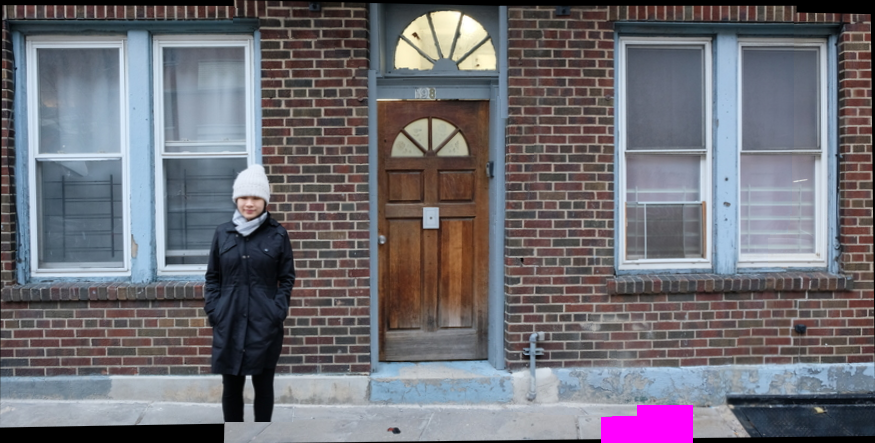}{1.0}
				\caption{\label{fig:3_image_ours_noblend} Our result with occlusion detected}
			\end{subfigure}
			\begin{subfigure}[t]{.5\textwidth}
				\centering
				\includegraphics[width=\textwidth, valign=t]{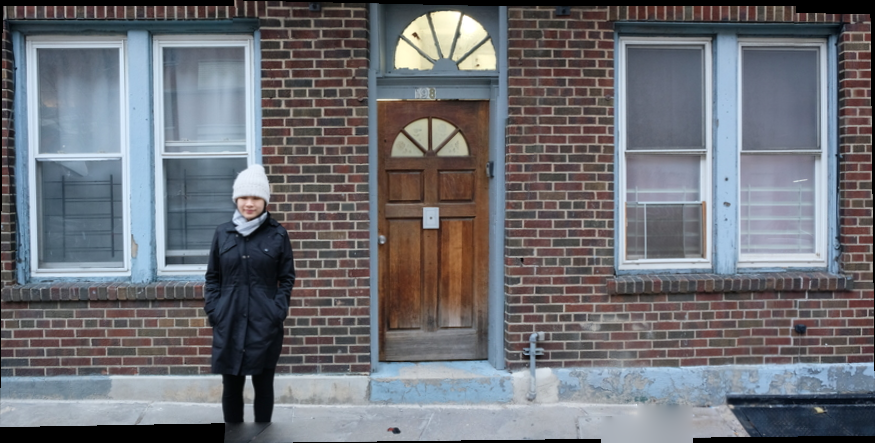}
				\caption{\label{fig:3_image_ours_blend} Our blend result}
			\end{subfigure}	
			\begin{subfigure}[t]{.5\textwidth}
				\centering
				\includegraphics[width=\textwidth, valign=t]{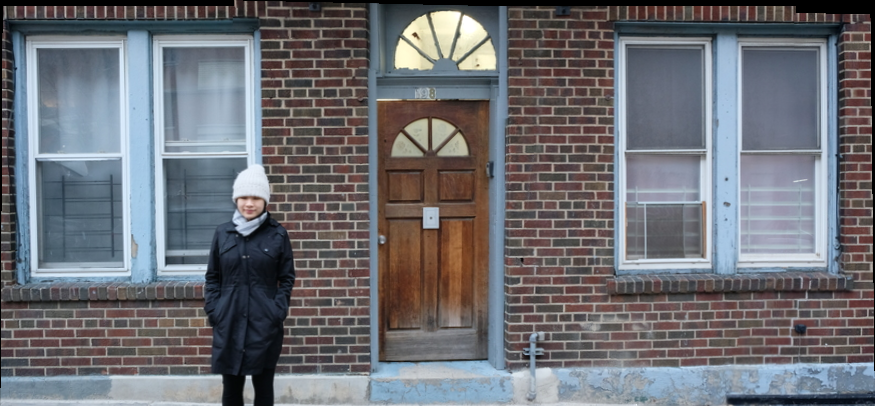}
				\vspace{6px}
				\caption{\label{fig:3_image_ours_cropped} Our cropped result}
			\end{subfigure}
			\caption{\label{fig:3-image} Three image stitching. In \ref{fig:3_image_ours_noblend}, we choose to not use the human in the right-most input. However, the legs block any information regarding the sidewalk, making this location occluded. Our algorithm correctly labels it as occluded and colors it magenta. \ref{fig:3_image_ours_blend} and \ref{fig:3_image_ours_cropped} present ways to fix this occlusion.}
		\end{minipage}
	\end{figure}

	\section{Conclusions, limitations and future work}
	\label{sec:limits}
	
	We have demonstrated that object detectors can be used to avoid a large class of visually jarring image stitching errors. Our techniques lead to more realistic and visually pleasing outputs, even in hard problems with perspective changes and differences in object motion, and avoid artifacts such as object duplication, cropping, and omission that arise with other approaches. Additionally, object detectors yield ways of evaluating the output of stitching algorithms without any dependence on the methods used.
	
	One potential drawback to our approach is that it applies only
	to inputs containing detectable objects, and provides no benefit in e.g. natural scenes where current object
	detection techniques are unable to generate accurate bounding boxes for elements such as mountains or
	rivers. We expect, however, that our techniques will become increasingly useful as object detection and scene
	matching improve. At the other end of the spectrum, we may be unable
	to find a seam in inputs with a large number of detected
	objects. We note that our crop, duplication, and omission terms are all soft
	constraints. In addition, objects can be prioritized based on saliency measures
	or category (i.e. human vs. other), and crops penalized more highly for objects deemed important. One existing
	use case where this might apply is for city imagery with pedestrians moving on sidewalks, 
	such as the content of Google Streetview. Traditional seam finding techniques
	find this setting particularly difficult, and torn or duplicated
	humans are easily identifiable errors.
	
	False positives from object correspondences are another issue. In this
	case, matching thresholds can be adjusted to obtain the desired
	behavior for the particular use case. Scenes with a large number of
	identical objects, such as traffic cones or similar cars, present a
	challenge when correspondence techniques are unable to match the objects to one another by taking advantage of
	the spatial characteristics of the input images. One issue that our technique cannot
	account for is identical objects with different motions: a
	pathological example might be pictures of identically-posed twins wearing the same clothing. We consider these
	false positives to be a reasonable tradeoff for improved performance in the more common use case.

	\subsubsection{Acknowledgements.} This research has been supported by NSF grants IIS-1161860 and IIS-1447473 and by a Google Faculty Research Award. We also thank Connie Choi for help collecting images.
	
	\ignore{
		\subsection{Comparison with our multiple registration approach}
		
		In our parallel submission \cite{Paper1} we take an alternative
		approach to improving stitching by relying on \emph{multiple
			registrations}. While there is some overlap in the classes of images
		that the two approaches are able to handle, there are also large
		differences. The object-centered approach discussed here uses a single
		registration, and fails when there are a large number of perspectives
		that must be integrated by considering multiple registrations. In
		contrast, the multiple registrations approach has no safeguards
		against errors that require structural error terms such as the ones
		introduced here to detect. We have found different instances where
		exactly one of the two approaches succeeds and the other fails, which
		we include in our supplemental material.

		\subsubsection{Conclusions}
		A natural direction to explore is to combine the ideas from the two
		approaches and use both multiple registrations and object terms in a
		single stitching approach. We have experimented with this strategy but
		encountered certain technical difficulties. Overcoming them to get the
		best of both worlds remains a challenge for future work.
	}

	\clearpage
	\newpage
	{\small
		\bibliographystyle{splncs}
		\bibliography{ourbib}
	}
	
\end{document}